\documentclass[10pt,cleanfoot]{asme2ej}

\usepackage{amssymb}
\usepackage{epstopdf}
\usepackage{etoolbox}
\usepackage{cite}
\usepackage{picinpar}
\usepackage{amsmath}               
  {

  }
\usepackage{url}
\usepackage{flushend}
\usepackage{textcomp}
\usepackage{bm}
\usepackage{comment}
\usepackage{xcolor}%
\usepackage{soul}%
\usepackage{tikz}
\newcommand*{\tikzhl}[1]{\tikz[baseline=(X.base)] \node[fill=lightgray] (X) {#1};}
\usepackage{makecell}
\UseRawInputEncoding
\usepackage{graphicx} 
\usepackage{epsfig}
\usepackage{fancyhdr}
\usepackage{setspace}
\usepackage{helvet}
\usepackage{hyperref}   
\hypersetup{
	colorlinks=true,
	linkcolor=blue,
	citecolor=blue,
	urlcolor=blue,
}
\usepackage[square,numbers]{natbib}

\title{Physics-data hybrid dynamic model of a multi-axis manipulator for sensorless dexterous manipulation and high-performance motion planning }

\author{Wu-Te Yang
    \affiliation{
	Department of Mechanical Engineering\\
	National Taiwan University\\
        Taipei, Taiwan 10617\\
        Email: wtyang@berkeley.edu
    }	
}

\author{Jyun-Ming Liao
    \affiliation{
        Department of Mechanical Engineering\\
	National Taiwan University\\
        Taipei, Taiwan 10617\\
        Email: r07522818@ntu.edu.tw
    }
}

\author{Pei-Chun Lin*
    \affiliation{
	Department of Mechanical Engineering\\
	National Taiwan University\\
        Taipei, Taiwan 10617\\
        Email: peichunlin@ntu.edu.tw
        }
}

\begin{document}

\maketitle

\hfill
\begin{abstract}
{\it 
We report on the development of an implementable physics-data hybrid dynamic model for an articulated manipulator to plan and operate in various scenarios. Meanwhile, the physics-based and data-driven based dynamic models are studied in this research to select the best model for planning. The physics-based model is constructed using the Lagrangian method, and the loss terms include inertia loss, viscous loss, and friction loss. As for the data-driven model, three methods are explored, including DNN, LSTM, and XGBoost. Our modeling results demonstrate that, after comprehensive hyperparameter optimization, the XGBoost architecture outperforms DNN and LSTM in accurately representing manipulator dynamics. With approximately 500k training data points, the XGBoost model closely matches the performance of the physics-based model, as assessed by the Root Mean Square Error (RMSE) between actual and estimated manipulator torque. The hybrid model with physics-based and data-driven terms has the best performance among all models based on the same RMSE criteria, and it only needs about 24k of training data. In addition, we developed a virtual force sensor of a manipulator using the observed external torque derived from the dynamic model and designed a motion planner through the physics-data hybrid dynamic model.The external torque contributes to forces and torque on the end effector, facilitating interaction with the surroundings, while the internal torque governs manipulator motion dynamics and compensates for internal losses. By estimating external torque via the difference between measured joint torque and internal losses, we implement a sensorless control strategy which demonstrated through a peg-in-hole task. Lastly, a learning-based motion planner based on the hybrid dynamic model assists in planning time-efficient trajectories for the manipulator. This comprehensive approach underscores the efficacy of integrating physics-based and data-driven models for advanced manipulator control and planning in industrial environments.}
\end{abstract}

\clearpage

\section{Introduction}\label{sec1}

\indent Industrial automation has been developed for the past 60 years for various tasks; for example, inspection~\cite{r1}, machining~\cite{r2,r3}, and pick-and-place~\cite{r4}. Along with high speed, high precision, and low cost, machine intelligence is regarded as one of the key performance indices of robots. Recently, during the fourth industrial revolution (i.e., Industry 4.0), the focus has been on large-scale smart factories, which rely on many factors, such as the Internet of things, cyber-physics systems, big data, and artificial intelligence \cite{c1}. All of these require a dynamic model of the machine as a crucial and basic building block. Without an appropriate model, the behaviors of the machine cannot be predicted planned, and controlled effectively. 

The dynamic model of the machine includes various motion effects such as inertia forces, Coriolis forces, gravitational force, and some losses. Traditionally, the model was developed based on the physics principle and phenomenon of the system, using either the Newton-Euler method or the Lagrangian method. However, the model of a complex system, such as a manipulator, is incapable of covering all the physic behaviors of the system, especially nonlinear behaviors, subtle unmodeled dynamics, and manipulator specification variations due to manufacturing processes. For example, the joint of the manipulator may have loss terms other than (Coulomb) friction and (viscous) damping, but in ordinary modelling work, we usually only consider these two because they have the most significant effects. The presence of manipulator specification variations due to manufacturing processes indicates that the commercial manipulator’s specifications are not the same as those listed in the datasheet, such as the D-H table. In this case, a data-driven model is quite useful due to its experience-based learning of the phenomenon and the behavior of the manipulator. 

In this work, following our initial exploration of developing a physical-based model of manipulators \cite{c2}, we extends our exploration to reliable data-driven and hybrid models (i.e., to cover un-modeled loss terms) that were derived using machine learning \cite{c3, c4}, where three methods were explored: deep neural network (DNN) \cite{c5}, long short-term memory (LSTM) \cite{c6}, and decision tree \cite{c7}. For example, Gillespie et al. \cite{c8} and Boucetta et al. \cite{c9} used DNN to deal with the uncertainty of the dynamic model for flexible and soft robots. In contrast, LSTM, a type of recurrent neural network (RNN), is commonly used for situations with time series. For example, Lai et al. \cite{c10} described a way to solve the problem of adaptive fuzzy inverse compensation control for uncertain nonlinear systems with generalized dead-zone nonlinearities in uncertain actuators . Thus, the generalized dead zone of the motor has hysteresis, where the values of friction in forward and reverse rotation are different. Without proper suppression, the nonlinear dead-zone may cause more errors and even instability of the system. Therefore, some researchers have tested whether LSTM can make the fitted model more accurate \cite{c6, c11}. Finally, the decision tree approach uses a tree model to determine the consequences of events. The most famous method is XGBoost \cite{c12}. It has been widely recognized in many machine learning and data mining challenges, such as Kaggle and the KDD Cup \cite{c13}, which shows the potential to handle un-modeling dynamics of robotic systems. 

In addition to the dynamic model, a smart manipulator requires the use of many sensors \cite{c14, c15}. Force/torque sensors are a popular choice and have led to many applications that require contact interactions, such as assembly tasks and human-robot collaborations \cite{c16, c17}. For example, Peg-in-hole is a common assembly task in factories. Various works have addressed this issue. Lin \cite{c18} developed a plug-in system for large transformers using a 6-axis robotic arm. Kim et al. \cite{c19} developed a hole-detection algorithm for square peg-in-hole applications using force-based shape recognition. Tang et al. \cite{c20} developed an autonomous alignment strategy for peg-in-hole. Li et al. \cite{c21} utilized a multi-sensor perception strategy to enhance the autonomy of uncertain peg-in-hole tasks. Beltran-Hernandez et al. \cite{c22} developed a variable compliance-control strategy for robotic peg-in-hole assembly. These three works utilized force/torque sensors. Due to the high cost of multi-axis force/torque sensors, other strategies for performing the peg-in-hole task have been proposed. For example, Park et al. \cite{c23} developed a low-cost peg-in-hole assembly strategy using contact compliance and without using a physical force/torque sensor. At present, most researches on virtual force sensors focus on the safety of human-machine collaboration (such as Yen et al. \cite{c24} and Li et al. \cite{c25}). Similar applications often require the establishment of mathematical models and the establishment of compensation for friction to achieve more accurate accuracy. For example, Lai et al. \cite{c10} developed a method to compensate the dead zone of the general actuator through the fuzzy adaptive inverse compensation method. This research hopes to establish all models at once through machine learning. Considering the high demand of the force/torque sensing information on the end-effector of a manipulator, virtual force/torque transducers seem to be a worthwhile research topic \cite{c24, c26} and motivated us to initially explore this aspect in this paper. One of the popular methods to develop a virtual force/torque sensor of a manipulator is to estimate the force/torque values by subtracting internal torques from measured torques. Estimating the internal torque of the manipulator requires a high-precision dynamic model, so this extension smoothly follows the development of dynamic models reported in this work.

Except for dexterous manipulation, which required force/torque estimations or measurements, trajectory planning is another important issue. Traditionally, the trajectory planning of industrial robots relies on physics models \cite{c31}. However, with the advancements in machine learning, the typical trajectory optimization can be achieved by AI algorithms. For example, Tian \cite{c27} and Stevo \cite{c28} demonstrated the application of genetic algorithms to optimize trajectories for robot arms efficiently.  Furthermore, reinforcement learning (RL) algorithms \cite{c29} have emerged as another popular method for trajectory optimization. Stulp et al. \cite{c30} successfully implemented RL techniques to generate optimal motions for a robot manipulator, particularly in pick-and-place tasks. The RL algorithms aim to learn optimal behaviors in an environment given user defined policy~\cite{RL}. Thus, the integration of a reliable hybrid dynamic model of a robot arm and RL algorithms forms a trajectory planner, promising enhanced efficiency and adaptability for industrial robots in various tasks.

In short, motivated by the emerging development of data-driven approaches, we reported on developing a reliable physics-data hybrid dynamic model for an industrial robot. The performance investigation of the physics-based and data-driven dynamic models, and the possible hybrid physics-data dynamic models is conducted. While the reported research usually focuses on using a specific modeling method for a specific application, we are interested in finding a reliable and general performance trend of the dynamic model from the aspect of its composition. In addition, following the core development of the data-driven dynamic model of the manipulator, we reported our initial investigation of the manipulation applications and a motion planner using the hybrid dynamic model. The contributions of this work include the following:
\begin{enumerate}
    \item[$\bullet$] Developing an implementable physics-data hybrid dynamic model of the articulated manipulator among various data-driven dynamic models using machine learning techniques, such as DNN, LSTM, and XGBoost and evaluating their performance. 
    \item[$\bullet$] Proposing an external torque observer based on the developed data-driven models and then proposing a virtual force/torque sensor via the observed external torque for robotic manipulation.
    \item[$\bullet$] Proposing a sensorless peg-in-hole assembly strategy inspired by human operation.
    \item[$\bullet$] Developing a learning-based time-efficient motion planner based on the physics-data hybrid dynamic model and is validated by experiments.
\end{enumerate}

To position our contributions among existing literature, we compare our approach with others. Reinhart et al. \cite{c32} studied data-driven forward and inverse dynamic models of an industrial robot for pure feedforward control. Polverini et al. \cite{c33} applied a physics-data hybrid dynamic model of a robotic system for tracking controller design. By contrast, we attempt to find an accutate digital twin of the robot arm for sensorless dexterous manipulation and high-performance motion planning. Yu et al. \cite{c34} developed a data-driven dynamic model for a fish robot to handle fluid dynamic issues; however, this research aims to compensate for the un-modeled dynamics of an articular manipulator. Xu et al. \cite{c35} built a data-driven dynamic model for a cable-driven planar robot to predict its tension and collision conditions. Nevertheless, our data-driven model is not only used to plan motions for a robot arm but also applied to realize a torque observer for sensorless manipulation and learning-based motion planning. Lastly, our previous work \cite{c2} proposed a learning-based motion planner, but we implemented a more accurate digital twin with the hybrid dynamic model and adjusted the policy for RL to generalize the previous framework. Overall, a functional physics-data hybrid dynamic model is constructed for sensorless dexterous manipulation and learning-based motion planning of an industrial robot in various industrial applications. 

The remainder of this paper is organized as follows. Section~\ref{sec2} introduces the dynamic models, and Section~\ref{sec3} describes the performance of the dynamic models. Section~\ref{sec4} describes the design of a virtual force sensor. Section~\ref{sec5} displays the design of the motion planner. Section~\ref{sec6} reports the experimental results and Section~\ref{sec7} concludes this research.

\section{Dynamic Models}\label{sec2}

The dynamic models utilized in this work are developed using two different approaches; one is a physics-based dynamic model, and the other is a data-driven model. The former is intrinsic, which allows us to truly understand the dynamics of the manipulator, but un-modeled dynamics, such as nonlinear terms, are difficult to identify. Furthermore, if the model has time-dependent terms (i.e., after a long operation time, the friction and damping terms are usually different), a remodel of the system is necessary. In contrast, the data-driven model is better able to capture nonlinear terms, but the trade-off is that the data collection and learning processes may take considerable time. While we are aiming to use a complete data-driven model for the tasks in this work, we are also interested in exploring the performance and dynamic characteristics of the physics-based model and the physics-data hybrid model for comparison purposes. Thus, this section describes the construction of the physics-based model and data-driven model separately.

\subsection{The physics-based models}\label{subsec1}
The physics-based dynamic model of the system is derived using the Lagrangian method. By forming the Lagrangian (L=T-V) of the manipulator, which is composed of kinetic energy (T) and potential energy (V), the equation of motion (EOM) of the system can be derived according to the Lagrangian equation.
\begin{align}
\frac{d}{dt}(\frac{\partial}{\partial \dot{q_i}}) - \frac{\partial}{\partial {q_i}} = {Q_i}', i = 1,...,n
\label{eqn: 1}
\end{align}
where ${q_i}$ and ${Q_i}'$ are the i-th generalized coordinate and the i-th generalized non-conservative force, respectively. The symbol $n$ represents the system's independent degrees of freedom (DOF). After derivation and rearrangement of the terms, the equations of motion (EOM) of the system can generally be expressed as
\begin{align}
M(q)\ddot q + C(q,\dot q)\dot q + G(q) = Q'
\label{eqn: 2}
\end{align}
where $M(\cdot)$, $C(\cdot)$, and $G(\cdot)$ represent the inertia term, Coriolis and centrifugal term, and gravitational term, respectively.

\begin{figure}[http]
    \centering
    \includegraphics[width=300pt]{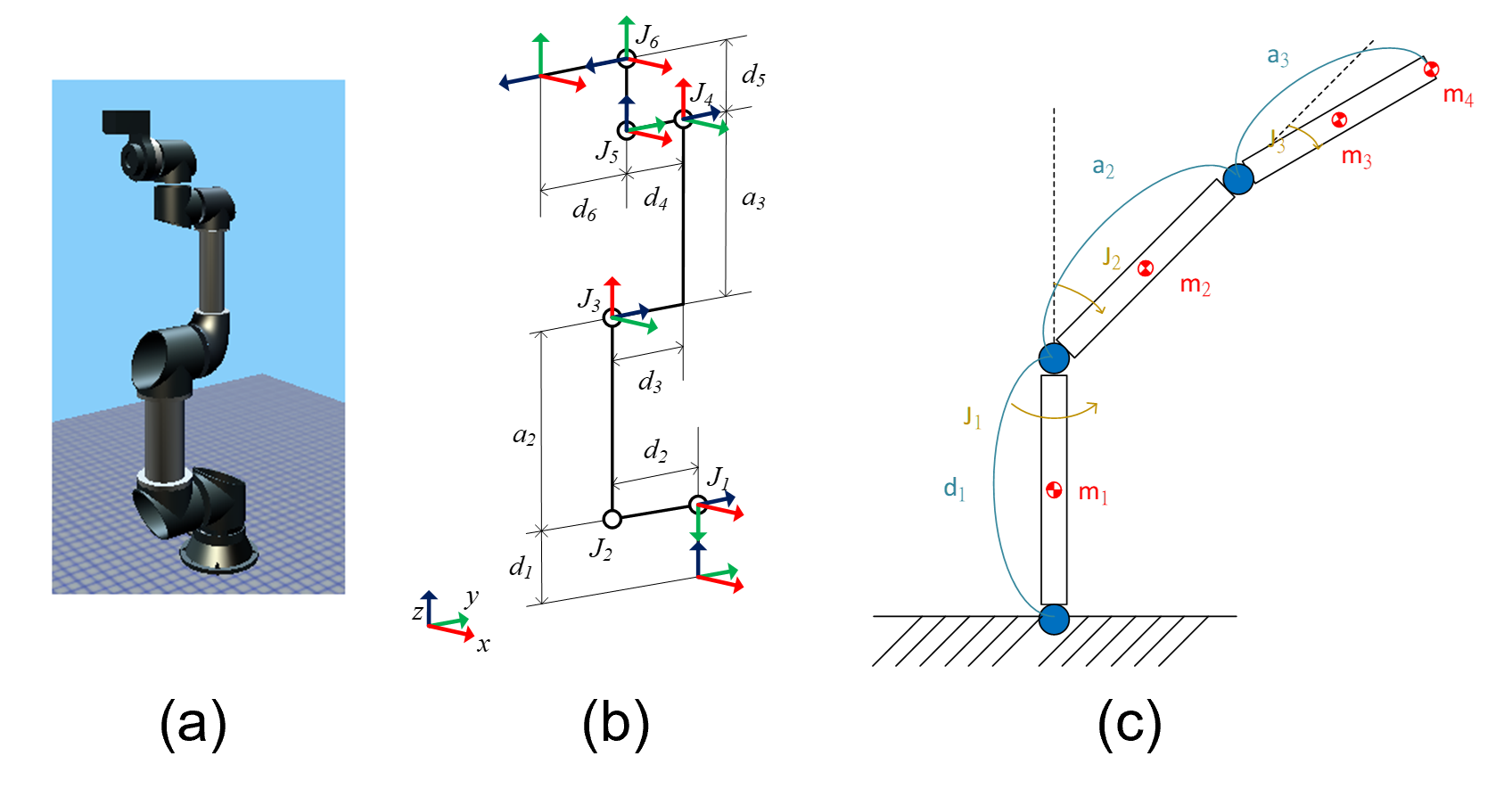}
    \caption{The manipulator TM5-700: (a) The CAD model, (b) the configuration, and (c) the simplified dynamic model.}
    \label{fig: 1}
\end{figure}

In this work, the derivation of EOMs is carried out for a 6-DOF articulated and collaborative manipulator (TM5-700, Techman Robot Inc.). The CAD model and configuration of the manipulator are shown in Figure~\ref{fig: 1}. This robot is utilized as the experimental testbed for the proposed methodology. Using the joint angles of the manipulator $(\theta)$ as the generalized coordinate, the general form of the physics-based EOM of the manipulator can be expressed as 
\begin{align}
EOM_{physics}(\theta, \dot \theta, \ddot \theta) = M(\theta)\ddot \theta + C(\theta,\dot \theta)\dot \theta + G(\theta) = Q'
\label{eqn: 3}
\end{align}
For the ideal manipulator without loss, the non-conservative force $Q'$ only contains joint actuation torque $(\tau_{motor})$:
\begin{align}
EOM_{physics}(\theta, \dot \theta, \ddot \theta) = M(\theta)\ddot \theta + C(\theta,\dot \theta)\dot \theta + G(\theta) = Q' = \tau_{motor}
\label{eqn: 4}
\end{align}
The joints are controlled to move according to the defined motion, either position-controlled (i.e., angle), velocity-controlled (i.e., angular speed), or force-controlled (i.e., torque). This ideal physics-based model of the manipulator shown in (\ref{eqn: 3})-(\ref{eqn: 4}) is hereafter referred to as Model P1. 

In contrast, the empirical manipulator generally has a loss, so the non-conservative force $Q'$, in general, contains two parts:
\begin{align}
EOM_{physics}(\theta, \dot \theta, \ddot \theta) = M(\theta)\ddot \theta + C(\theta,\dot \theta)\dot \theta + G(\theta) = Q' = \tau_{motor} - \tau_{loss}
\label{eqn: 5}
\end{align}
This realistic physics-based model of the manipulator is hereafter referred to as Model P2. The $\tau_{loss}$ is assumed to be generated at the joints, including inertia loss $(B_m \ddot \theta)$, viscous loss $(C_m \dot \theta)$, and friction loss $(f_c (sign(\dot \theta)))$:
\begin{align}
\tau_{loss}(\theta, \dot \theta, \ddot \theta) = B_m \ddot \theta + C_m \dot \theta + f_c (sign(\dot \theta))
\label{eqn: 6}
\end{align}
where $f_c$ and $sign(\cdot)$ represent the magnitude and direction of the friction force, respectively. No loss is assumed in other parts of the manipulator. Note that equations (\ref{eqn: 4}) and (\ref{eqn: 5}) represent different scenarios we have tested. Equation (\ref{eqn: 4}) indicates that the only non-conservative force of the manipulator model $(Q')$ is motor torque $(\tau_{motor})$ (i.e., Model P1), and equation (\ref{eqn: 5}) indicates that the manipulator model also includes the loss terms $(\tau_{motor}-\tau_{loss})$ (Model P2). Because the models contain different terms, their performance is expected to be different. In addition, note that the terms of $\ddot \theta$ in (\ref{eqn: 3}) and (\ref{eqn: 6}) are different. The $M(\theta) \ddot \theta$ in (\ref{eqn: 3}) represents inertia forces and the $B_m \ddot \theta$ represents loss proportional to the acceleration $\ddot \theta$.

\subsection{The data-driven Models}\label{subsec2}

In addition to constructing the model using physics principles, the model can also be data-driven. Following a similar form as the $EOM$ shown in (\ref{eqn: 2}), the data-driven $EOM$ can be expressed as 
\begin{align}
EOM_{data}(\theta, \dot \theta, \ddot \theta) = \tau_{motor}
\label{eqn: 7}
\end{align}
where the right side of the equation is the active joint torque $(\tau_{motor})$, which is utilized to support various effects of the manipulator’s dynamic motion, including inertia, Coriolis and centrifugal forces, gravitational forces, and losses. More specifically, the left side of (\ref{eqn: 7}) contains the effects $f_1 (\theta, \dot \theta, \ddot \theta)$ shown in ($\ref{eqn: 3}$) and $f_2 (\theta, \dot \theta, \ddot \theta)$ shown in (\ref{eqn: 6}). Because the exact form of the $EOM$ is unknown, the abstract function $f(\cdot)$ is utilized to express the resultant effects. In this formulation, the nonlinear effects can be included, which are difficult to cover explicitly in the physics-based formulation. Note that the manipulator may also contain some dynamic effects outside the input states $(\theta, \dot \theta, \ddot \theta)$, such as jerk dynamics, which need $\dddot \theta$. These effects are ignored in the modeling work.

Equation (\ref{eqn: 7}) is utilized as the data-driven $EOM$, where the manipulator states $(\theta, \dot \theta, \ddot \theta)$ in the left side and the torque $\tau_{motor}$ in the right side are utilized as the input (i.e., feature) and output states (i.e., label) for machine learning, respectively. From the aspect of physics, when the manipulator moves at the states of $(\theta, \dot \theta, \ddot \theta)$, the joints of the manipulator at this specific instant should supply the joint torque $\tau_{motor}$ if the model is precise. Therefore, it would be easy to evaluate the performance of the model by comparing the difference between its estimated $\tau_{motor- estimated}$ and actual manipulator torque $\tau_{motor}$. The dataset used for model training was experimentally generated using TM5-700, which will be described in detail in Section~\ref{sec3}.

The complete dynamic model contains 6 DOFs, and the training process is divided into subgroups to reduce the model’s complexity and the required dataset size. The 6-DOF articulated manipulator is generally designed to use the first three axes for translational motion (i.e., joints 1-3) and the last three axes for rotational motion (i.e., joints 4-6). The last three axes are compactly designed to have their rotational axis orthogonally intersected with each other, so Pieper’s solution can be deployed to algebraically or geometrically solve the inverse kinematics problem of the manipulator. Because of orthogonality, joints 4 and 5 are trained individually, yet the position states of other joints are imported so that the gravitational effects can be correctly considered. In contrast, the first three axes have a broad motion range, and their motion dynamics are highly coupled especially the second and third axes. Therefore, the first three axes are trained together. Furthermore, because the motion of the last three axes is comparably small compared to that of the first three axes, in the training process of the first three axes, the motion of the last three axes is ignored and could be regarded as a small point mass; $m_4$ is the sum of the last three axes’ mass, mounted at the end of manipulator as shown in Figure~\ref{fig: 1}(c). The machine learning models of axes 1, 2, and 3 are established separately from the models of axes 4 and 5. We tried to train the model by including all axes of the manipulator as inputs. However, the model's performance is not promising, even when the data number is doubled. Because the first three axes are for the translational motion of the manipulator, the effects of the axes are coupled and should be trained together. Axes 4 and 5 are designed for rotational motion within a small range, so their effect can be separated from the first three axes to include training efficiency.

\begin{figure}[http]
    \centering
    \includegraphics[width=300pt]{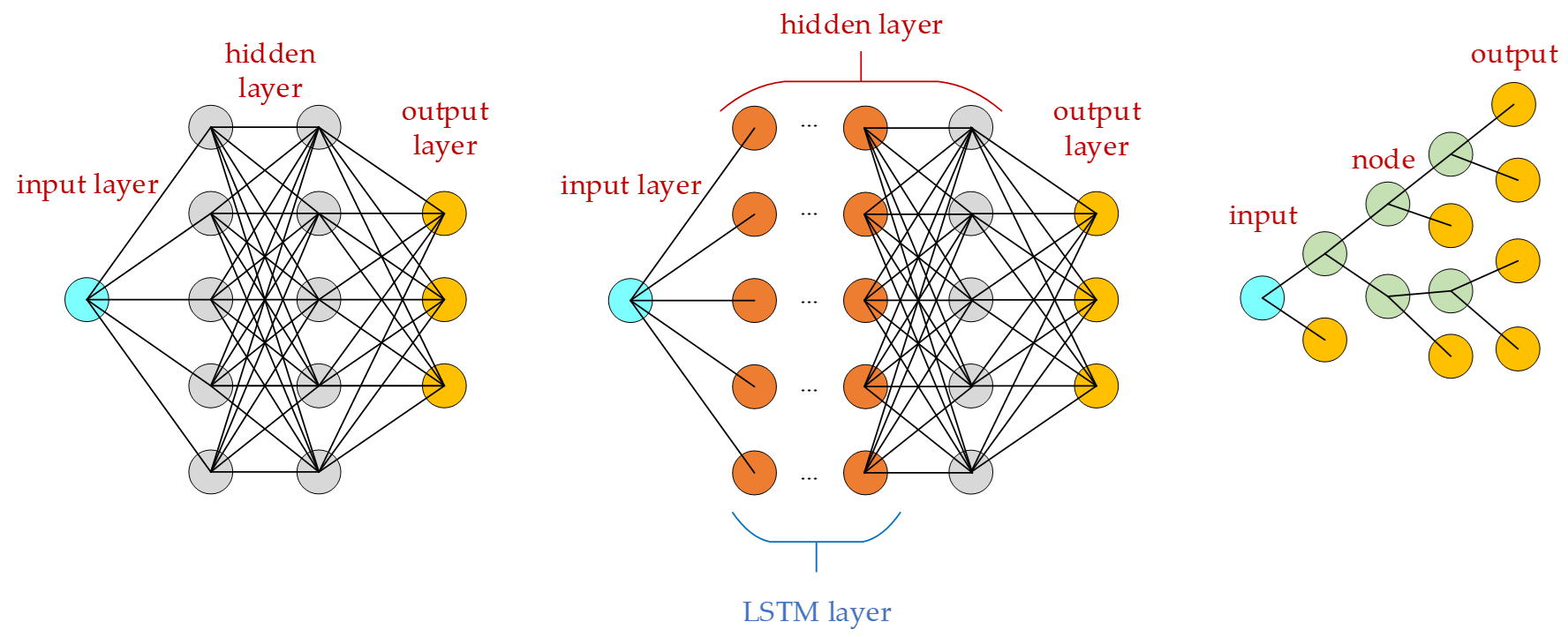}
    \caption{Structural diagrams of (a) DNN, (b) LSTM, and (c) XGBoost. The cyan and yellow circles represent the input and output of the model, respectively. In the case of XGBoost, shown in (c), only one of the outputs is the actual output, depending on the result of the final judgment of the input value in the decision tree.}
    \label{fig: 2}
\end{figure}

As described in the introduction, three methods (DNN, LSTM, and XGboost) are explored to evaluate their performance in constructing a data-driven dynamic model of the manipulator, whose structure diagrams are shown in Figure~\ref{fig: 2}. The neuron (orange circle in the figure) of LSTM is different from DNN. LSTM layers whose cell (i.e., orange circles in the figure) is memorable and composed of an input gate, a cell state, forget, and output gates~\cite{c11}. XGBoost (Extreme Gradient Boosting) is a Gradient Boosted Tree (GBDT). Each time the original model is kept unchanged, and a new function is added to the model to correct the error of the previous tree to improve the overall model. The structure shown in the far right side of Figure~\ref{fig: 2} is a set of classification and regression trees (CART). Each leaf of the regression tree corresponds to a set of values, which are used as the basis for subsequent classification. The yellow part in the figure may be the output of the XGBoost model, which depends on the result of the final judgment of the input value in the Decision Tree to determine which yellow circle is the final output~\cite{c12}. Note that the utilized DNN package (keras) and XGBoost package (XGBoost) have different architecture~\cite{c36, c37}, where the former supports multiple outputs, but the latter does not. Therefore, the three estimations $(\tau_1, \tau_2, \tau_3)$ of the model using XGBoost needed to be trained separately. Table~\ref{tab: Table 1} lists the features and labels of the training process. The training of joint 6 was skipped since it was not utilized in the following applications.

\begin{table}[http]
\centering
\caption{\label{tab: Table 1}The features and labels of the manipulator used for training.}
    \begin{tabular}{|c c|} 
    \hline
    The translational axis (joint 1-3) &\\ [0.8ex] 
    \hline
    Feature $(rad, rad/s, rad/s^2)$ & $\theta_1$, $\dot \theta_1$, $\ddot \theta_1$, $\theta_2$, $\dot \theta_2$, $\ddot \theta_2$, $\theta_3$, $\dot \theta_3$, $\ddot \theta_3$  \\ [0.8ex] 
    Label (N-m) & $\tau_1$, $\tau_2$, $\tau_3$ \\[0.8ex] 
    \hline
    The rotational axis (joint 4-6) &\\[0.8ex] 
    \hline
    Feature $(rad, rad/s, rad/s^2)$ & $\theta_1$, $\theta_2$, $\theta_3$, $\theta_4$, $\theta_5$, $\theta_6$, $\dot \theta_j$, $\ddot \theta_j$, $j = 4,5$ \\ [0.8ex] 
    Label (N-m) & $\tau_j$, $j = 4,5$ \\[0.8ex] 
    \hline
    \end{tabular}
\end{table}

\begin{figure}[http]
    \centering
    \includegraphics[width=300pt]{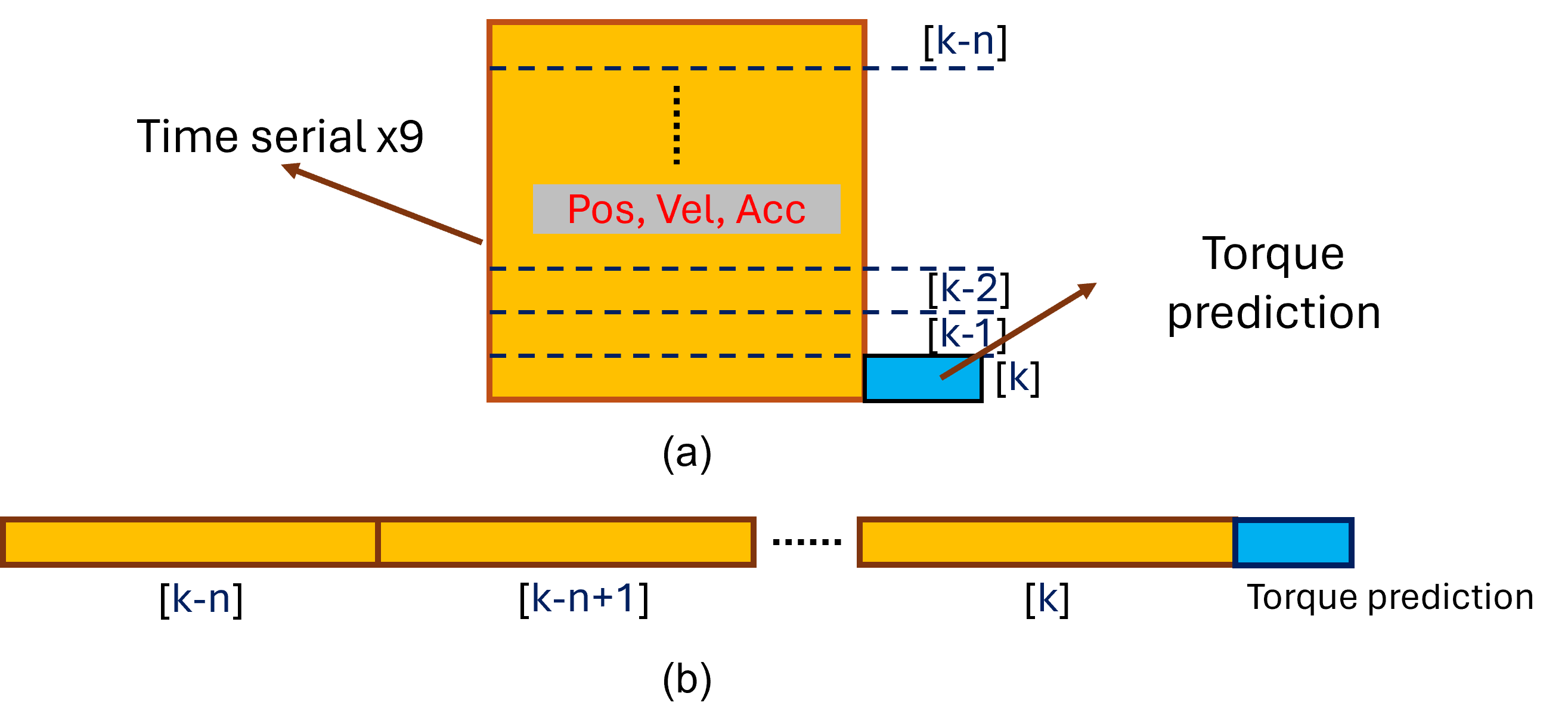}
    \caption{The inputs/features and outputs/labels of the model using the LSTM method (a) and XGBoost method (b).}
    \label{fig: 3}
\end{figure}

Unlike the DNN, which uses states of one moment to predict the states of the next moment, the LSTM uses time sequence as input data, and the XGBoost can be modified in the same manner. The structure of LSTM replaces neurons in the original hidden layer of the DNN model with LSTM units; thus, the underlying learning strategy is completely different. The LSTM is prone to overfitting, so the architecture is more complicated than the previous DNN, and dropout and ${recurrent\_dropout}$ must be considered. While the states of a series of timestamps are fed into the model, the input becomes 2-dimensional as shown in Figure~\ref{fig: 3}(a). For any predicted torque at any timestamp [k] shown in the blue block, the model requires the states at timestamp [k-n] to [k]. In our implementation using the LSTM method, $n=9$ is utilized. The sampling period is 8 ms during the experiment, so the model used the data of the past 80 ms, shown in yellow blocks, as information for prediction. In the case of the first three axes, the number of states input into the model is 90 (i.e., 9 states multiplied by 10 timestamps). The model using XGBoost is also trained with time sequence data. Because XGBoost could not accept array inputs, the data needed to be flattened into a one-dimensional array as shown in Figure~\ref{fig: 3}(b). For these models with sequential data, the abstracted form of the data-driven model shown in (\ref{eqn: 7}) is still applicable, but the required inputs $(\theta, \dot \theta, \ddot \theta)$ contain states at several different timestamps.

After describing the model construction, as well as its features and labels, the following paragraphs in this section will describe the selection of hyperparameters, which need to be set before training. If we use LSTM as an example, the hyperparameters include several factors, such as the learning rate, number of iterations, number of layers, and number of neurons in each layer. The choice of hyperparameters seriously affects the capability and performance of the model, but the selection of high-quality hyperparameters is very complicated. Thus, a methodical search strategy is crucial~\cite{hyper,c38,c39}. Four methods are commonly used to adjust hyperparameters: manual, grid search, random search, and Bayesian optimization. The manual method mainly relies on a variety of different architectures to test and train individually and finally adjust the hyperparameters according to the result of the loss function or objective function. The grid search method sequentially defines the values to be searched according to the type of hyperparameter. The program runs through all the hyperparameter combinations to test and select the best parameters based on the loss function or objective function, which is time-consuming because every cartesian product of a hyperparameter combination must be tried. Although this brute-force method is time-consuming, it can find the best solution as long as the exhaustive hyperparameters are sufficiently comprehensive. The random search method is similar to the grid search method. The difference is that it randomly selects a combination of parameters to search. Relatively speaking, it may be more economical in time and computing resources than the grid search method, but the search may be insufficient. Finally, the essence of Bayesian optimization lies in the verification of previous results and the probability of selecting hyperparameters for the next iteration~\cite{c40}.

This work used two methods to adjust the hyperparameters of machine learning. One is grid search due to its complete search in the parameter space, and the other is Bayesian optimization due to its efficiency. The work then compared the performance of the manipulator models using these two adjusting methods. GridSearchCV in the scikit-learn module is utilized~\cite{c41}. For a DNN model with multiple hidden layers and a large amount of training data, a grid search is time-consuming and impossible to run simultaneously. For example, a model with 10 hyperparameters, each with nine variations, would have to run $9^{10}$ trials to find the best hyperparameter. Even comparing the search results would be time-consuming. Therefore, an assisted lookup table is used to instantly update the best hyperparameter values during the search process, as shown in \textbf{Algorithm 1}. Thus, the grid search could independently adjust the respective hyperparameters individually, so only $9\times10$ attempts are required. Since changes in the neurons and activation functions of the previously hidden layer affect the next hidden layer, it is necessary to repeatedly adjust the parameters to confirm their consistency. Therefore, assuming that the process needs to be repeated three times, the attempts become 9×10×3. In this work, the 492,038 data points (80 \% for the training set and 20 \% for the validation set) are utilized, and the model is set to have a seven-layer hidden layer, each with a set of neuron numbers and activation functions. Therefore, there are 14 hyperparameters to adjust in total, and it took about 2–3 days to execute using an ordinary desktop computer equipped with an Intel i7-6800k CPU, NVIDIA RTX 2080 GPU, and 16 GB of RAM. Note that the comprehensiveness of the hyperparameter space determines the quality of the searched hyperparameter.

\begin{table}[http]
\centering
    \begin{tabular}{|l|}
    \hline
    \textbf{Algorithm 1.} Automatic independent hyperparameter grid search \\
    \hline
    \textbf{Take NN with two hidden layers as an example (with Python scikit-learn GridSearchCV).}\\
    Assuming that you only want to adjust part of the hidden layer, the initial value is defined as follows.\\
    \textbf{Initial parameter definition:}\\
    \underline{Record hyperparameter:}\\
    i. HyperParaList = ['neu1','acti1', 'neu2','acti2','optimizer','lr'].\\
    ii. HyperParaValue = [64, 'linear', 64, 'linear',' Adam',0.001].\\
    \underline{Hyperparameter value lookup table:}\\
    i. neuron = [64, 128,256,512,1024,2048].\\
    ii. activation = ['softplus','softsign','relu','tanh','selu','linear','elu'].\\
    iii. optimizer = ['SGD','Adagrad','Adadelta','Adam','Adamax','Nadam'].\\
    iv.	lr = [0.0005, 0.001, 0.0015, 0.002, 0.0025, 0.003].\\
    \textbf{Output:}
    Every time a single hyperparameter adjustment is completed, the best value recorded\\ by HyperParaValue is printed and saved to a CSV file:\\
    Ex.: HyperParaValue:[128, 'tanh', 128, 'tanh', 'Adam', 0.0005].\\
    \textbf{Procedure:}\\
    \textbf{Step 1.}Read the first value of the list HyperParaList and HyperParaValue\\ as the \underline{GridSearchCV} to adjust the hyperparameters.\\
    The first value of HyperParaList is 'neu1' and its type is a neuron, so \underline{GridSearchCV} will use\\ neuron as the search table. Other hyperparameter values are defined according to HyperParaValue.\\
    \textbf{Step 2.} After the \underline{GridSearchCV} search is completed, the best value will overwrite\\ 
    the position in HyperParaValue corresponding to the HyperParaList search.\\
    \textbf{Step 3.} Repeat steps 1 and 2 until the end of the list to complete a complete NN\\
    hyperparameter search.\\
    \textbf{Step 4.} Then, you can search again from the beginning based on the results of this \\
    complete search to ensure that the hyperparameters are stable.\\
    \hline
    \end{tabular}
\end{table}

The Bayesian optimization search method, in contrast, establishes a probability model through the corresponding relationship between the loss function results obtained by the model with the previously generated hyperparameters. Thus, each time the hyperparameters are adjusted, a probability model is attempted to estimate the minimum loss function, where the adjustment of the hyperparameters is executed more efficiently~\cite{c36, c38}. The objective function of the Bayesian optimization used in this work is L2 loss or mean square error (MSE):
\begin{align}
MSE = \frac{1}{m} \sum_{k=1}^{m} y_k - \hat{y_k} \
\label{eqn: 8}
\end{align}
As the value of the objective function decreases, the better the solution.

\subsection{The physics-data hybrid models}\label{subsec3}

While either the physics-based model or the data-driven model can be separately utilized to model the manipulator’s dynamic behaviors, these two models can also be combined, forming the so-called physics-data hybrid model. In this method, the data-driven part acts as the compensation term to cover the un-modeled dynamics ignored in the physics-based part. This is advantageous because the inclusion of the data-driven part increases the model accuracy, yet the required data size is not as large as the pure data-driven model which requires a large data set to capture the complex dynamic behavior of the manipulator.

Various compositions could be utilized to form the hybrid models. The first trial in this work involved taking all the physics-based terms from (\ref{eqn: 3}) and (\ref{eqn: 6}) into account and using a data-driven model $(f_{h1}(\theta, \dot \theta, \ddot \theta))$ to compensate for un-modeled effects; this model is hereafter referred to as Model H1:
\begin{align}
EOM_{physics}(\theta, \dot \theta, \ddot \theta) + \tau_{loss}(\theta, \dot \theta, \ddot \theta) + f_{h1}(\theta, \dot \theta, \ddot \theta) = \tau_{motor}
\label{eqn: 9}
\end{align}
The second trial considered motion dynamics without loss modeling and used a data-driven model $(f_{h2}(\theta, \dot \theta, \ddot \theta))$ to model the loss and compensate for un-modeled effects; this model is hereafter referred to as Model H2:
\begin{align}
EOM_{physics}(\theta, \dot \theta, \ddot \theta) + f_{h2}(\theta, \dot \theta, \ddot \theta) = \tau_{motor}
\label{eqn: 10}
\end{align}
In contrast, the third trial keeps the loss part and used a data-driven model $(f_{h3}(\theta, \dot \theta, \ddot \theta))$ to model motion dynamics and other model effects; this model is hereafter referred to as Model H3:
\begin{align}
\tau_{loss}(\theta, \dot \theta, \ddot \theta) + f_{h3}(\theta, \dot \theta, \ddot \theta) = \tau_{motor}
\label{eqn: 11}
\end{align}
The models constructed in this section are summarized in Table~\ref{tab: Table 2}. The data-driven only model is referred to as Model D1, which utilizes DNN-based architecture.

\begin{table}[http]
\centering
\caption{\label{tab: Table 2}Composition of various models with physics-based and/or data-driven portions.}
    \begin{tabular}{|c c c c|} 
    \hline
    Model & $f_{h1}(\theta, \dot \theta, \ddot \theta)=$ & $f_{h2}(\theta, \dot \theta, \ddot \theta)=$ & Data-driven model\\ [0.5ex] 
    {} & $M(\theta)\ddot \theta + C(\theta, \dot \theta)\dot \theta$ & $B_m\ddot \theta + C_m\dot \theta$ &{} \\ [0.5ex]
    {} & $+ G(\theta)$ & $ + f_c({sign(\dot \theta)})$ & {} \\[0.5ex]
    \hline
    Model P1 & V & {} & {} \\ [0.5ex]
    Model P2 & V & V & {} \\ [0.5ex]
    Model D1 & {} & {} & V \\[0.5ex]
    Model H1 & V & V & V \\[0.5ex]
    Model H2 & V & {} & V \\[0.5ex] 
    Model H3 & {} & V & V \\[0.5ex]
    \hline
    \end{tabular}
\end{table}

\subsection{Motion data generation of the manipulator}

Data collection is one of the key elements of the machine learning process, and its quality strongly determines the performance of the learned model. This is especially true for complex dynamic systems, which have many states coupled with each other. The manipulator in this work is a typical complex dynamic system, whose complete global training data are difficult to collect~\cite{c42}. Coupled states are especially difficult to collect because they include not only position but also velocity and acceleration data. While data diversity is crucial, efficiently collecting a sufficient amount of data with broad diversity is very important. In this section, the trajectory generation method, which quickly generates diversified training data, is described.

Motion data generation of the manipulator includes three steps. First, the motion range of each joint of the manipulator is digitized into a selected angular interval, so the original infinite possible joint angles are transformed into finite angular configurations. For example, if all six joints of the manipulator are digitized into $N$ segments, there exist $(N+1)^6$ manipulator configurations. Second, removing the configuration causes a collision. Third, trajectories of the manipulator are generated by permutation of the possible manipulator configurations. For example, if the manipulator has M configurations, there exists $P_2^M$ possible trajectories. The trajectories are organized into a list, where the end point of one trajectory matches the starting point of the next trajectory. Thus, $P_2^M$ trajectories are arranged into one continuously movable trajectory. For each trajectory, the manipulator is set to move at V different speeds. From the original manipulator configurations, there are $P_2^M V$ generated trajectories.

\begin{figure}[http]
    \centering
    \includegraphics[width=90pt]{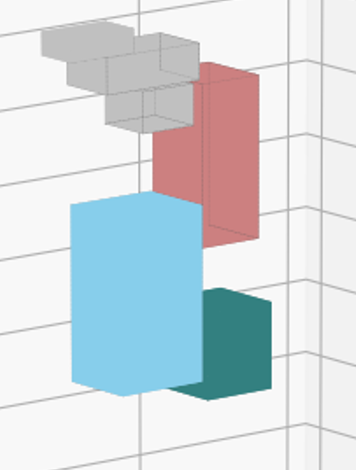}
    \caption{The combination of simplified bounding-box model of the manipulator and hybrid dynamic model of the TM5-700 as a simulator.}
    \label{fig: 4}
\end{figure}

The empirical implementation of trajectory generation on TM5-700 is described as follows. First, the rotation ranges of the first to the sixth joints are $\pm 270^o, \pm 80^o, \pm 150^o, \pm 180^o, \pm 180^o$, and $\pm 270^o$, respectively. In practical applications, different joints use a variety of angle values, depending on the motion range of the manipulator. The values also include both ends of the motion range, so the whole rotation range is covered. Increasing the number of joint angles of all axes will increase the coverage of the manipulator workspace. Note that both the position and the speed of the joint should be varied. Therefore, the position and speed combination will increase dramatically if more joint angles are set. We consider this work to be our first trial, so the number of joint angles is set at values feasible for training and experiments. Second, the manipulator is modeled using rectangular orientation bounding boxes (OBB)~\cite{c43} as shown in Figure~\ref{fig: 4}, and then the model is utilized for a collision check using the separation axis ~\cite{c43} theorem (SAT)~\cite{c44}. The surrounding objects residing within the workspace of the manipulator are also modeled and checked for collision to ensure that the manipulator can be driven safely. After the collision check, four configurations are left, so there are six trajectories after permutation. Each trajectory is set to run at three different speeds, so 18 trajectories in total are generated for data collection.

\section{Performance of the Dynamic Models}\label{sec3}

\subsection{Performance comparison of physics-based, data-driven, and hybrid dynamic models}\label{subsec31}

The models listed in Table~\ref{tab: Table 2} are trained using the DNN-based data-driven model with empirical motion data of a TM5-900 manipulator. The features and label details are listed in Table~\ref{tab: Table 1}. The DNN architecture used a hidden layer with 7 layers. Each hidden layer used batch normalization, which speeds up training completion, reducing the need for dropouts, and increasing the likelihood of early stopping. The number of neurons in each layer is 512, 1024, 1024, 1024, 1024, 1024, 512. The activation function is $tanh$, the last layer is the output layer, the activation function is linear, the loss function is the average absolute error, the optimizer is $adam$, and the $validation\_split$ = 0.2, which indicates that $80 \%$ and $20 \%$ of the data were training and verification data, respectively.

The evaluation is executed using the motion of the first three axes of the manipulator (i.e., joints 1\textendash3), which are highly coupled and dynamic. A total of 23,598 training data points are collected and shuffled before feeding them to the models. In addition to the training and verification data, test data are generated using six randomly selected manipulator motion trajectories, where each runs five times. Thus, there 90 joint trajectories in total are used for testing. For each trajectory, the root mean square error (RMSE) between the true torque generated by the manipulator $(\tau_{motor real})$ and the torque estimated by the models $(\tau_{motor})$ was computed: 
\begin{align}
RMSE = \sqrt{\frac{1}{N}\sum_{n=1}^{N} (\tau_{motor,i} - \tau_{motorreal,i})^2}
\label{eqn: 12}
\end{align}
where $N$ indicates the number of data points. Thus, for each model, there are 90 RMSE values. To evaluate the model statistically, the means, standard deviations, maximums, and minimums of the models are computed as the performance index of the model.

Table~\ref{tab: Table 3} lists the performance of the various models constructed in Section ~\ref{sec2}. The results reveal that the physics-based model without loss (i.e., P1) is not realistic and has large RMSEs. The added physics-based loss term greatly improves the model’s performance (i.e., P2). Furthermore, the added data-driven part further improves the model’s performance (i.e., H1), which indicates that the DNN captures the un-modeled dynamics of the manipulator. The hybrid models H2 and H3 separately used the data-driven model to compensate for the physics-based loss terms or the motion dynamics of the model, respectively. However, given the same dataset for training, the added unknown model dynamics decrease the performance of the models. H3 performing worse than H2 indicates that motion dynamics are more complex and difficult to learn than loss dynamics.

\begin{table}[http]
\centering
\caption{\label{tab: Table 3}Statistical performance of various models based on RMSEs between the torque generated by the manipulator and estimated by the models. (Unit$\colon N \cdot m)$.}
    \begin{tabular}{|c c c c c|} 
    \hline
    Model & mean & std & max & min \\ [0.5ex]
    \hline
    Model P1 & 9.554 & 3.297 & 18.171 & 3.754 \\
    \tikzhl{Model P2} & \tikzhl{3.990} & \tikzhl{0.955} & \tikzhl{6.316} & \tikzhl{2.256} \\
    Model D1 & 3.656 & 1.097 & 8.172 & 2.314 \\
    Model H1 & 5.440 & 1.332 & 7.501 & 2.393 \\
    Model H2 & 7.026 & 4.298 & 20.033 & 2.522 \\
    \tikzhl{Model H3} & \tikzhl{8.911} & \tikzhl{4.799} & \tikzhl{24.257} & \tikzhl{3.022} \\[0.5ex]
    \hline
    \end{tabular}
\end{table}

\subsection{Architecture and hyperparameter adjustment of the data-driven models}\label{subsec31}

Table~\ref{tab: Table 3} also reveals that with the same dataset, the pure data-driven model D1, which needs to learn all the dynamics of the manipulator, has the worst performance among all models with different compositions, which are listed in Table~\ref{tab: Table 2}. This result suggests that the amount of training data may not be inadequate, or the chosen hyperparameters are unsuitable for learning manipulator dynamics. Therefore, additional evaluations are executed to address these concerns. Table~\ref{tab: Table 4} shows the performance of the data-driven model with different training data and the same DNN architecture. The first two rows are identical to the data shown in Table~\ref{tab: Table 3} as the reference. The table clearly shows that the means of RMSE decreases when the amount of training data increases. The increase in training data effectively improves the performance of the data-driven model, but it is still not as good as the physics-based model P2.

Following the results shown in Table~\ref{tab: Table 4}, the hyperparameters of the DNN-based model are adjusted using a grid search and Bayesian optimization. Meanwhile, the variations of the model with different loss functions and layers are evaluated. The DNN-based models with hyperparameter adjustment are hereafter referred to as Model D2. In addition to the 7-hidden-layer architecture and L2 loss utilized in the model, the 3-hidden-layer architecture and L1 loss are evaluated, where the L1 loss represents the mean absolute error (MAE):
\begin{align}
MAE = \frac{1}{M}\sum_{n=1}^{M} |y_k - \hat{y_k}|
\label{eqn: 13}
\end{align}

\begin{table}[http]
\centering
\caption{\label{tab: Table 4}Statistical performance of the DNN-based models with different amounts of training data. (Unit$\colon N \cdot m)$.}
    \begin{tabular}{|c c c c c c|} 
    \hline
    Model & Training data & mean & std & max & min \\ [0.5ex]
    \hline
    Model P2 & {} & \tikzhl{3.990} & \tikzhl{0.955} & \tikzhl{6.316} & \tikzhl{2.256} \\
    {} & 23598 & 8.911 & 4.799 & 24.257 & 3.022 \\
    {} & 64840 & 8.330 & 3.416 & 16.068 & 3.063 \\
    Model D1 & 95717 & 7.141 & 2.492 & 12.100 & 3.547 \\
    {} & 331481 & 6.686 & 4.045	& 19.108 & 2.459 \\
    {} & \tikzhl{492038} & \tikzhl{6.093} & \tikzhl{2.281} & \tikzhl{10.764} & \tikzhl{2.483} \\[0.5ex]
    \hline
    \end{tabular}
\end{table}

\begin{table}[http]
\centering
\caption{\label{tab: Table 5}Statistical performance of the DNN-based models with different architectures and hyperparameters. (Unit$\colon N \cdot m)$.}
    \begin{tabular}{|c c c c c c c c c|} 
    \hline
    Model & {} & layers & hyperparameter & loss & mean & std & max & min \\ [0.5ex]
    {} & {} & {} & adjustment & {} & {} & {} & {} & {} \\ [0.5ex]
    \hline
    Model P2 & {} & {} & {} & RMSE & \tikzhl{3.990} & \tikzhl{0.955} & \tikzhl{6.316} & \tikzhl{2.256} \\
    Model D1 & DNN & 7 & {} & RMSE & 6.093 & 2.281 & 10.764 & 2.483 \\
    {} & \tikzhl{DNN} & \tikzhl{7} & \tikzhl{Grid search} & \tikzhl{MSE} & \tikzhl{4.702} & \tikzhl{1.249} & \tikzhl{8.065} & \tikzhl{2.446} \\
    {} & DNN & 7 & Grid search & MAE & 4.975 & 1.339 & 9.346 & 2.993 \\
    Model D2 & DNN & 3 & Grid search & MSE & 4.744 & 1.444 & 9.58 & 2.59 \\
    {} & DNN & 7 & {Bayesian optimization} & MSE & 4.777 & 1.416 & 8.341 & 2.506 \\[0.5ex]
    \hline
    \end{tabular}
\end{table}

Table~\ref{tab: Table 5} lists the results. A total of 492,038 training data points used are utilized. The first two rows are identical to the data shown in Table~\ref{tab: Table 4} as the reference. The last five rows represent the models with different numbers of hidden layers and loss functions, all with hyperparameter tuning by a grid search. Furthermore, \textbf{Algorithm 1} is utilized to speed up the tuning process. The last row represents the models whose hyperparameters are adjusted using Bayesian optimization.

Table~\ref{tab: Table 5} reveals that the mean values of RMSE of the D2 models whose hyperparameters are adjusted using grid search or Bayesian optimization of hyperparameters are similar, and they are all smaller than that of model D1. This indicates that the grid search adjustment of hyperparameters is very helpful in improving learning performance. The table also shows that the model with MSE loss is slightly better than the one with MAE loss. The table also reveals that the model with a 3-hidden layer can achieve similar performance as the model with a 7-hidden layer. In the current setup, the model using more than 3 hidden layers does not improve performance and just increases the amount of the training and tuning. The adjustment of hyperparameters using Bayesian optimization performs similarly to the grid search. However, the parameter search time is reduced from 48 hours for the grid search to 4 hours, so it is a very time-effective method. In addition to the reported models, we try many other DNN-based models with different settings, but the performance of the models seems limited. Therefore, other variations of the model are then explored.

\begin{table}[http]
\centering
\caption{\label{tab: Table 6}Statistical performance of the models. (Unit$\colon N \cdot m)$.}
    \begin{tabular}{|c c c c c c|} 
    \hline
    Model & Type & mean & std & max & min \\ [0.5ex]
    \hline
    Model P2 & Physics & 3.990 & 0.955 & 6.316 & 2.256\\
    Model D2 & DNN & 4.702 & 1.249 & 8.065 & 2.446\\
    Model D2 & LSTM & 4.151 & 1.673	& 9.139	& 2.205 \\
    Model D4 & XGBoost & 3.803 & 1.444 & 9.462 & 2.017\\[0.5ex]
    \hline
    \end{tabular}
\end{table}

The LSTM-based model takes a sequence of historical data as the input, which intuitively helps to capture the dynamics of the manipulator. The utilized LSTM model is a two-layer LSTM, where Gaussian noise is stuffed between the two parts and the last two fully connected layers. The LSTM model is referred to as model D3. Table~\ref{tab: Table 6} shows the results, where the first two rows are identical to the data shown in Table~\ref{tab: Table 5} as the reference. The table shows that the LSTM model (i.e., model D3) can indeed reduce the overall averaged RMSE, which confirms that time-serial features can improve machine learning performance. However, although the LSTM model performs better than the DNN model, it is still not as good as the physics-based model. Therefore, data-driven models with different architectures and principles are then explored. 

The XGBoost method is developed and its performance is compared using the DNN and LSTM models. The architecture of the decision-tree models is relatively simple, so the hyperparameter search is much easier than the NN method. There is no need to define each hidden layer, select an order arrangement, or, for example, decide whether to use Dropout or Batch Normalization or which activation function to use. The XGBoost model in this work included the adjustment of the following ten hyperparameters: $learning\_rate$, $max\_depth$, $min\_child\_weight$, $colsample\_bytree$, $subsample$, $reg\_alpha$, $reg\_lambda$, $gamma$, $n\_estimators$, and $seed$. The XGBoost model is hereafter referred to as model D4. Table~\ref{tab: Table 6} shows that although the RMSE standard deviation and RMSE maximum of the XGBoost model are relatively larger than those of the physics-based model, the RMSE means are smaller than those of the latter. This indicates that the pure data-driven model can effectively model the complicated dynamic motion of the manipulator.

In summary, the results reported in this section led to the following conclusions. First, with about 500k of training data, the XGBoost model (i.e., model D4) performs similarly to the physics-based model with both motion dynamic terms and loss terms (i.e., model P2), based on the judgment of RMSE between the actual manipulator torque and the estimated one using the model. Second, the hybrid model with physics-based and data-driven terms (i.e., model H1) has the best performance among all models based on the same RMSE criteria, and it only needs about 24k of training data. From the aspect of training data collection and training time, this is a very effective method to model the manipulator dynamics when compared to the pure data-driven model. The construction of the physics-based model of the manipulator is not trivial, either. Third, based on the amount of training data used for testing, increasing the amount used improves of performance, which matches the general observation of the learning model. Fourth, the adjustment of hyperparameters also improves the model’s performance. Bayesian optimization performs similarly to the grid search methods but requires much less time. Fifth, given the same dataset, the LSTM with sequential data performs better than the DNN with instant data, and XGBoost outperforms the LSTM model.

Note that many data-driven models are evaluated in this work. The fundamental idea is that we would like to determine which data-driven model is suitable for modeling ordinary differential equations (ODEs). Once a suitable model is found, the work can be easily extended to other systems that move, as all movable systems obey Newton’s Second Law and can be expressed as second-order ODE systems. However, because movable systems appear in many different forms or compositions, only some of them have physics-based models constructed. If a new system is designed, especially a high DOF system, it is also challenging to construct its physics-based model. Therefore, we think that the data-driven approach has merit and is worth trying.

\section{Design of a Virtual Force Sensor}\label{sec4}

The states of the end effector and its force interaction with the environment are important information for manipulator operation. While the states of the end effector can easily be observed using the manipulator joint configuration with forward kinematics, the force interaction is difficult to obtain, as it includes three forces and three sources of torque. One general approach is to install a 6-axis force sensor on the end effector. However, this greatly increases the hardware cost of the manipulator~\cite{c24}. Following the modeling work presented in previous sections, the estimated joint torque of the manipulator using machine learning is further developed into the estimation of interaction forces between the end effector and the environment. This includes two steps: estimating the required torque at the joints, which are utilized to generate the required forces at the end effector, and establishing the mapping between the former and the latter.

\subsection{The external force observer}\label{subsec41}

As described in~(\ref{eqn: 5}), if the manipulator moves freely within the workspace with states $(\theta, \dot \theta, \ddot \theta)$, the joints are required to generate $\tau_{motor}$ to support this motion. Here, $\tau_{motor}$ is rewritten as $\tau_{motor\_free}$ to make the presentation clearer:  
\begin{align}
EOM_{physics}(\theta, \dot \theta, \ddot \theta) = M(\theta)\ddot \theta + C(\theta, \dot \theta)\dot \theta + G(\theta) = Q' = \tau_{motor\_free} - \tau_{loss}
\label{eqn: 14}
\end{align}

When the manipulator has some payload or has a force interaction with the environment, the torque generated at the joints needs to be adjusted:
\begin{align}
\tau_{motor} = \tau_{motor\_free} - \tau_{ext}
\label{eqn: 15}
\end{align}
More specifically, the partial torque of the total joint torque $(\tau_{motor})$ is transmitted to the end effector for force interaction. This torque is referred to as external torque $(\tau_{ext})$. The remaining torque is utilized to generate its dynamics and to compensate for the loss. Therefore, the external torque can be derived as
\begin{align}
\tau_{ext} = \tau_{motor} - \tau_{motor\_free} = \tau_{motor} - EOM_{physics}(\theta, \dot \theta, \ddot \theta) - \tau_{loss}
\label{eqn: 16}
\end{align}
Empirically, $\tau_{motor}$ is directly available from the motor drive board of the commercial manipulator utilized in this work (TM5-700, Techman Robot Inc.), so we just need to log the torque data when collecting the experimental data for model learning. The $\tau_{motor\_free}$ is estimated using the data-driven model developed in Sections \ref{sec2}-\ref{sec3}. 

\begin{figure}[http]
    \centering
    \includegraphics[width=320pt]{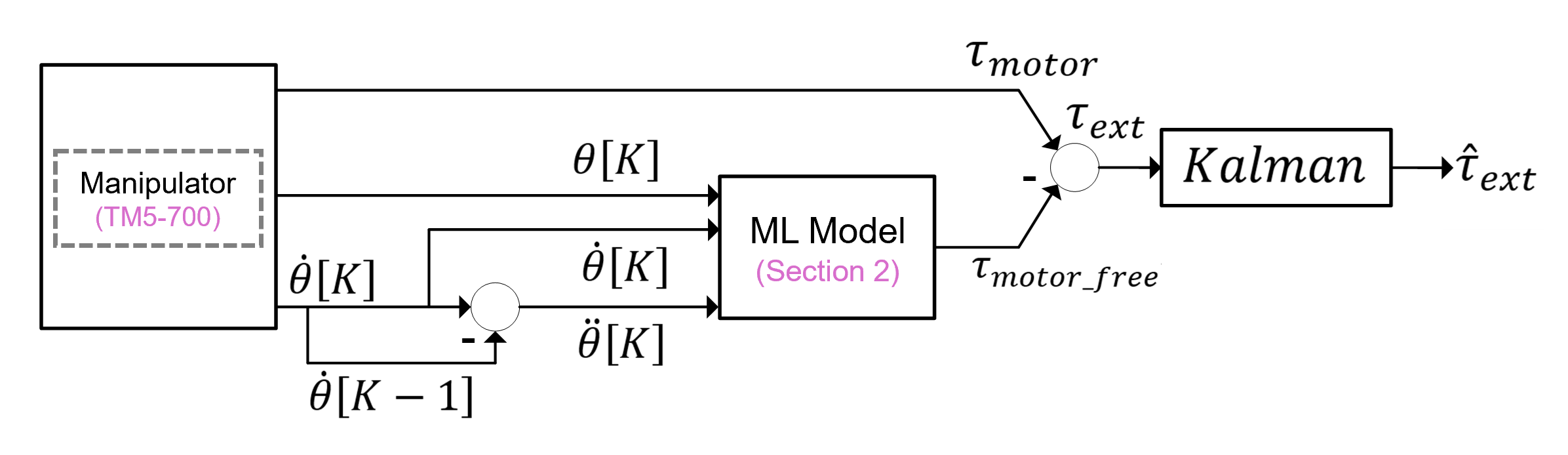}
    \caption{The architecture of the external torque observer.}
    \label{fig: 5}
\end{figure}

In the empirical implementation, because the external torque $\tau_{external}$ compute from $\tau_{motor}$ and $\tau_{motor\_free}$ is noisy, the Kalman filter (KF)~\cite{c45} is utilized:
\begin{align}
\hat\tau_{ext} = Kalman[\tau_{ext}]
\label{eqn: 17}
\end{align}
The KF system contained only one torque state without introducing other motion states. In the time update of the KF process, the prediction model is set to be the same as the previous value. The measurement update utilizes the computed torque $\tau_{ext}$ as shown in~(\ref{eqn: 16}). The measurement noise is obtained by analyzing the model noise and joint-torque noise, and the process noise is adjusted through actual experiments. The overall architecture of the external torque observer is shown in Figure~\ref{fig: 5}.

\subsection{Virtual force sensor}\label{subsec42}

The virtual force sensor uses XGBoost models, which has the best analysis effect above. Given the estimated external torque $\tau_{ext}$, the corresponding force/torque on the end effector could be estimated to serve as the virtual force sensor. Mapping from the external joint torque to the forces/torque at the end effector can be computed using the Jacobian in the force domain~\cite{c31}, which can also be regarded as a quasi-static estimation of the force flow in the manipulator. Following a similar strategy where the dynamic motion of the manipulator can be approximated using a data-driven model, the Jacobian, which is based on the kinematic relationship, should be able to be modeled using a data-driven model. Therefore, instead of deriving the Jacobian of the manipulator, the virtual force sensor of the manipulator is developed using a data-driven approach.
\begin{figure}[http]
    \centering
    \includegraphics[width=350pt]{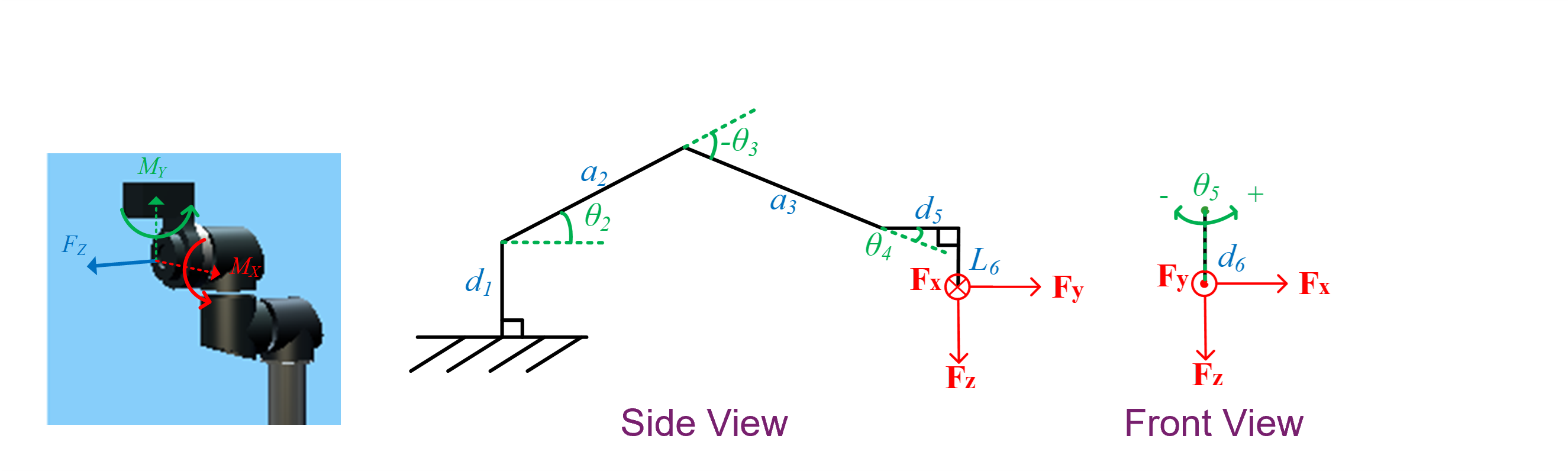}
    \caption{The 3-axis virtual force sensor developed in this work: (a) notations of the three estimated axes; (b) the major configuration of the manipulator when the sensor was utilized.}
    \label{fig: 6}
\end{figure}
In this work, a 3-axis virtual force sensor is developed using a data-driven model, including a normal force and two tilting moments of the end effector as shown in Figure~\ref{fig: 6}(a). While the manipulator is posed in certain configurations, the forces/torque on the end effector can be roughly approximated using a simpler static force relationship. This work utilizes a peg-in-hole application to demonstrate the virtual force sensor and the most common configuration of the manipulator in this application is shown in Figure~\ref{fig: 6}(b). In this configuration, the estimation of two moments $(M_X,M_Y)$ mainly rely on two external torques of the 4th and 5th joints $(\tau_{ext 4},\tau_{ext 5})$:
\begin{align}
M_X = f_{M_X}(\theta_1, \theta_2, \theta_3, \theta_4, \theta_5, \theta_6, \hat\tau_{ext 5})
\label{eqn: 18}
\end{align}
\begin{align}
M_Y = f_{M_Y}(\theta_1, \theta_2, \theta_3, \theta_4, \theta_5, \theta_6, \hat\tau_{ext 4})
\label{eqn: 19}
\end{align}
As for the normal force $F_Z$, it is modeled as
\begin{align}
F_Z = f_{F_Z}(\theta_1, \theta_2, \theta_3, \theta_4, \theta_5, \theta_6, \hat\tau_{ext 1}, \hat\tau_{ext 2}, \hat\tau_{ext 3})
\label{eqn: 20}
\end{align}
Considering the force flow of the manipulator, the forces on the end effector are propagated (or supported) by all the joints of the manipulator. Therefore, the torque at the first three axes is utilized to estimate $F_Z$, and that of the 4th and 5th joints is reserved to estimate $(M_X, M_Y)$. Similar to the model work in Section~\ref{sec2}-\ref{sec3}. These models utilize the XGBoost architecture, and the training data considers the inertia of the robotic arm, so the feature considers the angle, angular velocity, and angular acceleration of ten steps to predict the joint torque of the manipulator. 

\begin{figure}[http]
    \centering
    \includegraphics[width=100pt]{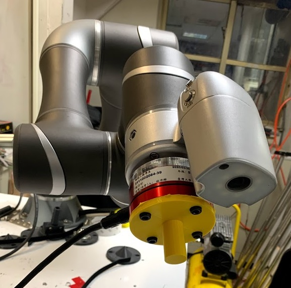}
    \caption{A commercial 6-axis force sensor (WEF-6A200-4-RCD, WACOH Inc.) is installed on the end effector to collect ground truth data for training and quantitative experiment validation.}
    \label{fig: 7}
\end{figure}

\begin{figure}[http]
    \centering
    \includegraphics[width=300pt]{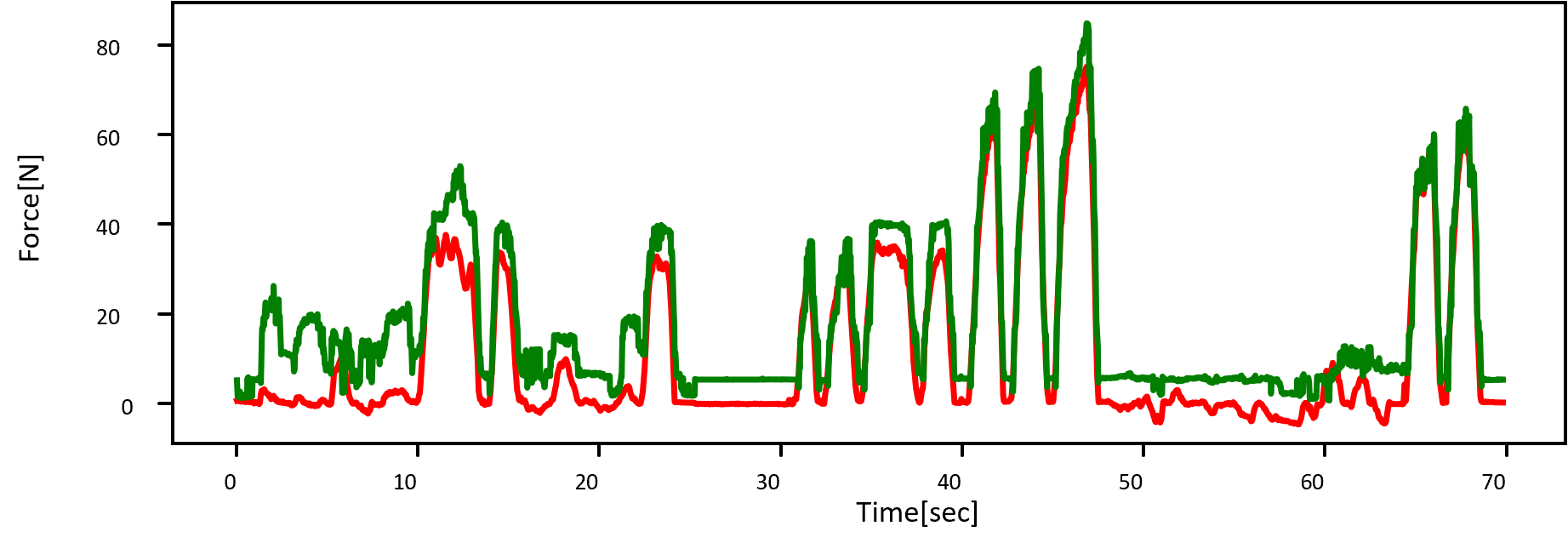}
    \caption{Time-series data of the estimated force (green dashed curve) and the measured force (red solid curves).}
    \label{fig: 8}
\end{figure}

To collect labeled data as the training dataset of the models shown in~(\ref{eqn: 18})-(\ref{eqn: 20}), a commercial 6-axis force sensor is installed on the end effector as the ground truth (WEF-6A200-4-RCD, WACOH Inc.), as shown in Figure~\ref{fig: 7}. During the training process, the end effector is posed in various configurations with forces/torque at different levels, and a total of more than 260,000 data points are collected for training. In the experiment, we try to select five postures to apply force on the force sensor and observe the force sensor values and results predicted by the model~(\ref{eqn: 20}). The force applies by the hand ranges from 0 to 75 N. The largest difference in force is mainly between 0 and 20 N. The maximum error is about 20 N, as shown in Figure~\ref{fig: 8}, and the average error is about 9.4 N. The maximum force error of 20N is due to the empirical system being affected by inertia, which is also supported by the physics-based models. Therefore, the estimation would be imprecise in the very beginning and then improve, as shown in Figure~\ref{fig: 8}. Furthermore, LSTM is a time-series model that uses a period of past data as the input, so the model takes time to converge for operation. The figure also shows that the model can capture the dramatic changes in the force, which indirectly indicates that the estimation system has sufficient bandwidth. However, the estimation of the force in a small magnitude is less accurate. We believe this phenomenon results from the manipulator’s mechanical properties (e.g., the dead zone of the joint, static friction, viscous damping forces), as well as the accuracy of the mechatronic systems (e.g., the accuracy of the current measurement).

\subsection{Data-driven models for the 4th and 5th joints of the manipulator}\label{subsec43}

\begin{figure}[http]
    \centering
    \includegraphics[width=300pt]{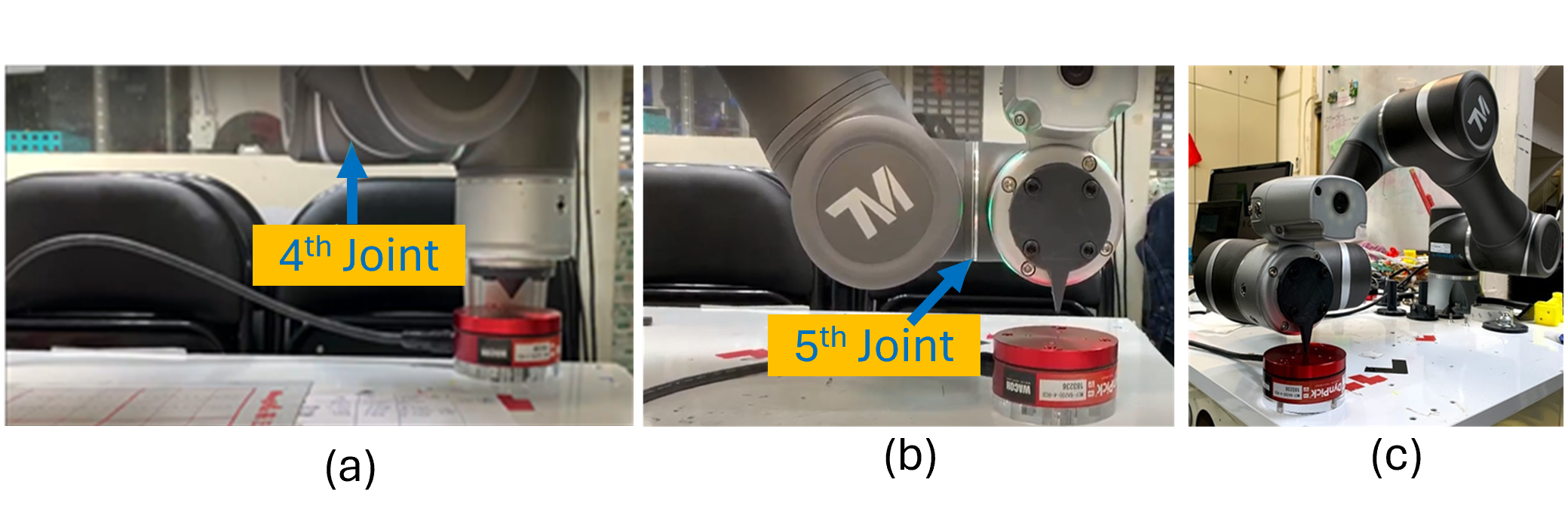}
    \caption{Experimental setup for validating the performance of the model for the 4th joint in (a) and the 5th joint in (b) and (c), which has two operating configurations.}
    \label{fig: 9}
\end{figure}

\begin{figure}[http]
    \centering
    \includegraphics[width=320pt]{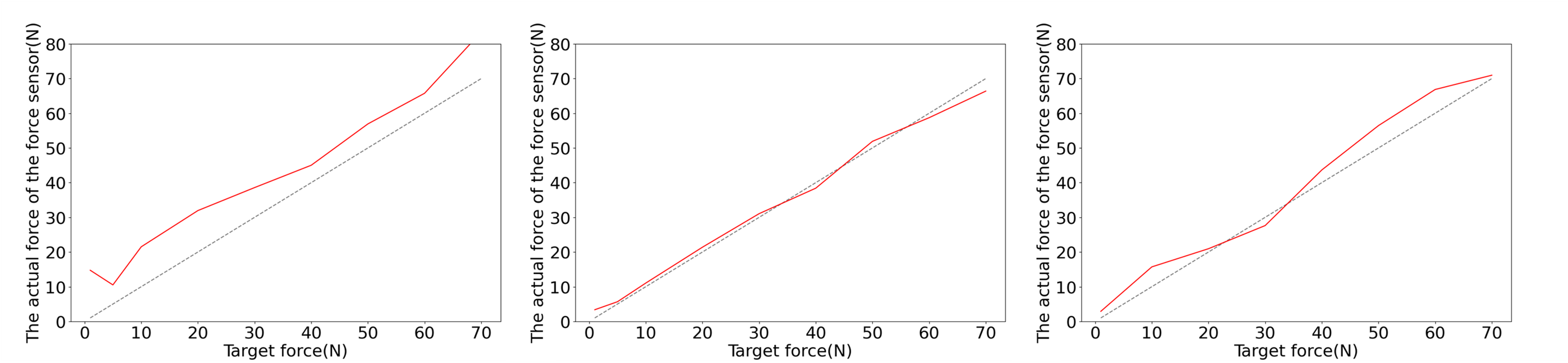}
    \caption{Comparison of the estimated force from the virtual force sensor (x-axis) and measured force (y-axis) from the 6-axis force sensor. The grey line represents the ideal condition.}
    \label{fig: 10}
\end{figure}

After describing the development of various models for the first three axes of the manipulator, this section focuses on the model development of the 4th and the 5th joints of the manipulator. As mentioned previously, due to orthogonality and the intersected joint axis, modeling of the 4th and the 5th joints is executed separately. In addition, because these joints of the manipulator are close to the end effector, instead of generating test data for evaluation, the performance of the joint torque estimation is directly observed using a multi-axis force sensor mounted on the end effector (WEF-6A200-4-RCD, WACOH Inc.), the ground truth. Figure~\ref{fig: 9} shows the manipulator layout for experimental validation of the 4th joint model and the 5th joint model, respectively. Joint 5 could be posed in two configurations as shown in Figure~\ref{fig: 9}(b)-(c). The joint torque is transformed to be represented as the force of the end effector using the method described in Section~\ref{subsec42}, so the measured data could be compared with that of the ground truth data. Note that the joint torque of the manipulator contributes to two tasks; external torque contributes to the forces/torque on the end effector, which interacts with the surroundings, and internal torque is utilized for manipulator motion dynamics and for compensating internal loss. Subtracting the internal torque (i.e., derived by the data-driven model) from the total joint torque (i.e., torque measured at the joints) yields the external torque, which is then transformed into the forces/torque at the end effector using Jacobian forces. The Jacobian defines the “instant” relationship between the joint torque and the forces/torque at the end effector, which, when viewed from the perspective of geometry, can be considered a virtual force with geometrical considerations. Figure~\ref{fig: 10} shows the estimated force from the virtual force sensor and the measured force from the 6-axis force. The RMSE of the estimated force in joint 4 and joint 5 experiments are $\pm2[N]$ and $\pm1[N]$, respectively. 

Table~\ref{tab: Table 7} lists the performance of the XGBoost model for the 5th joint. Here, the averaged error, standard deviation of the error, and maximum error between the real force measured by the multi-axis sensor and the force estimated by the model are utilized as the criteria. Four different settings were explored. Trial 1 utilizes instant input data (i.e., no sequential data). Trial 2 uses the same setup but with data normalization using
\begin{align}
(data - mean of data)  / std.of data
\label{eqn: 21}
\end{align}
Trial 3 utilizes sequential data inputs, where 10 timestamps are imported, the same number as the LSTM model. Trial 4 uses the same input data as Trial 3, and the data are normalized, similar to the process in Trial 2. The table shows that using sequential data helps to improve the performance of the XGBoost model, and data normalization is unnecessary. The general normalization method is to normalize the data to a distribution with a mean of 0 and a variance of 1. This is mainly for the effectiveness of the gradient, but here, we find that the normalization effect will not improve. The main reason may be that the original scaling value is compressed to between -1 and 1; thus, the resolution of the data value becomes smaller, so the effect is not improved. 

\begin{table}[http]
\centering
\caption{\label{tab: Table 7}Statistical performance of the models. (Unit$\colon N \cdot m)$.}
    \begin{tabular}{|c c c c|} 
    \hline
    Setting & $E_{max}$ & $E_{mean}$ & $E_{std}$ \\ [0.5ex]
    \hline
    1 & 29.469 & 6.754 & 3.844\\
    2 & 26.701 & 7.809 & 4.882\\
    3 & 20.654 & 4.320 & 3.115\\
    4 & 29.619 & 13.108 & 4.774\\[0.5ex]
    \hline
    \end{tabular}
\end{table}
\section{Design of a Motion Planner}\label{sec5}

In Section~\ref{sec3}, the accuracy and ability of the physics-data hybrid model to capture the dynamics of the articulated robot have been verified. Except for manipulation, an industrial robot relies on trajectory planning to pick and place objects. The physics-data hybrid model, therefore, is utilized to build a motion planner for the robot arm as the Figure~\ref{fig: 4}. The motion planner here focuses on speed optimization of the robot arm and aims to reduce the elapsed time of a chosen trajectory. The experimental results can be observed in Section~\ref{subsec63}

Deep reinforcement learning was employed for trajectory optimization due to its ability to streamline complex constraint settings and analytics while still yielding comparable results. The specific algorithm utilized in this context was proximal policy optimization (PPO), introduced by OpenAI in 2017~\cite{ppo}. PPO operates on an actor-critic architecture, where the actor determines actions such as adjusting the positions or timestamps of via points. Subsequently, the critic evaluates the actor's actions, assigning a score to guide the actor towards actions that lead to higher scores. In this scenario, the reward is designated for speed optimization, with trajectories receiving higher rewards for shorter elapsed times. The trajectory yielding the highest reward is considered the optimal one.

The motion planner here is to reduce the elapsed time from the start point, via the point, and end point of the trajectory. The reward function of RL is set as: 
\begin{equation} Rewards = 
    \begin{cases}
      a \cdot |\tau_{limit} - \tau_{peak}|, if~force~exceeds~the~limit\\
      b \cdot T_{reduction}, otherwise
    \end{cases}\,
    \label{eqn: RL}
\end{equation}
where $\tau_{limit}\in R^n$ is the limit of torque, $\tau_{peak}\in R^n$ is the peak torque of a joint of a robot, $n$ is the DOF of the planned robotic system, $T_{reduction}$ is the reduced elapsed time of a trajectory via RL optimization, a ($<$ 0) is a negative constant, and b ($>$ 100) is a positive constant.

This reward function~(\ref{eqn: RL}) attempts to penalize when torque exceeds the limit. The reward will be negative as the torque exceeds the limit. The more it exceeds the limit, the more negative rewards it obtains. The $T_{reduction}$ is a difference of elapsed time before and after optimization. That is, the PPO algorithm will implement the simulator based on the hybrid model to predict the dynamics of the manipulator and plan a speed-optimized trajectory via the given reward function. The optimized time-efficient trajectory will be generated and transferred to the manipulator for testing as discussed in Section~\ref{subsec63}.

\section{Experiment}\label{sec6}

The developed virtual force sensor described in Section~\ref{sec4} was evaluated using two experiments. The first one, wiping a table, involved moving the end effector of the manipulator on a flat surface with constant normal force, $F_Z$, estimated using the virtual force sensor. The second one involved performing the peg-in-hole task of the manipulator, which utilizes all three axes of the virtual sensor. To quantitatively evaluate performance, the same commercial force sensor shown in Figure~\ref{fig: 7} was utilized as the ground truth. The Kalman filter was applied to filter estimation noise. Additionally, a force control strategy was deployed to modulate the interaction forces between the manipulator and the objects.

\subsection{Wiping the table}\label{subsec61}

\begin{figure}[http]
    \centering
    \includegraphics[width=300pt]{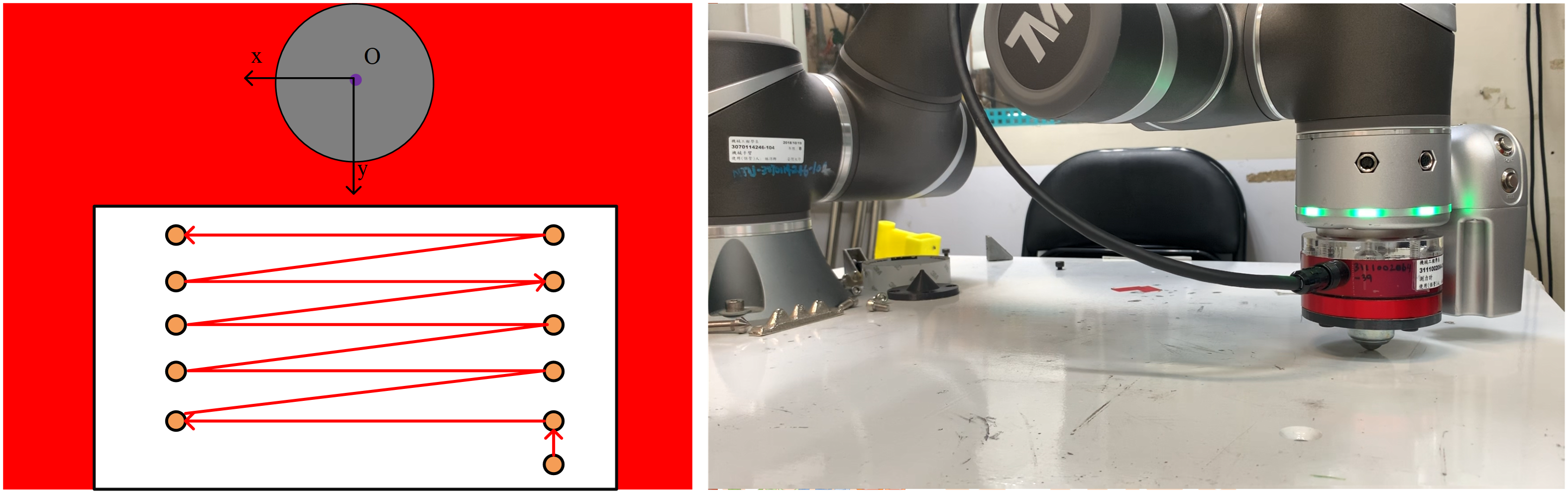}
    \caption{The trajectory and setup of the wiping the table experiment.}
    \label{fig: 11}
\end{figure}

\begin{figure}[http]
    \centering
    \includegraphics[width=320pt]{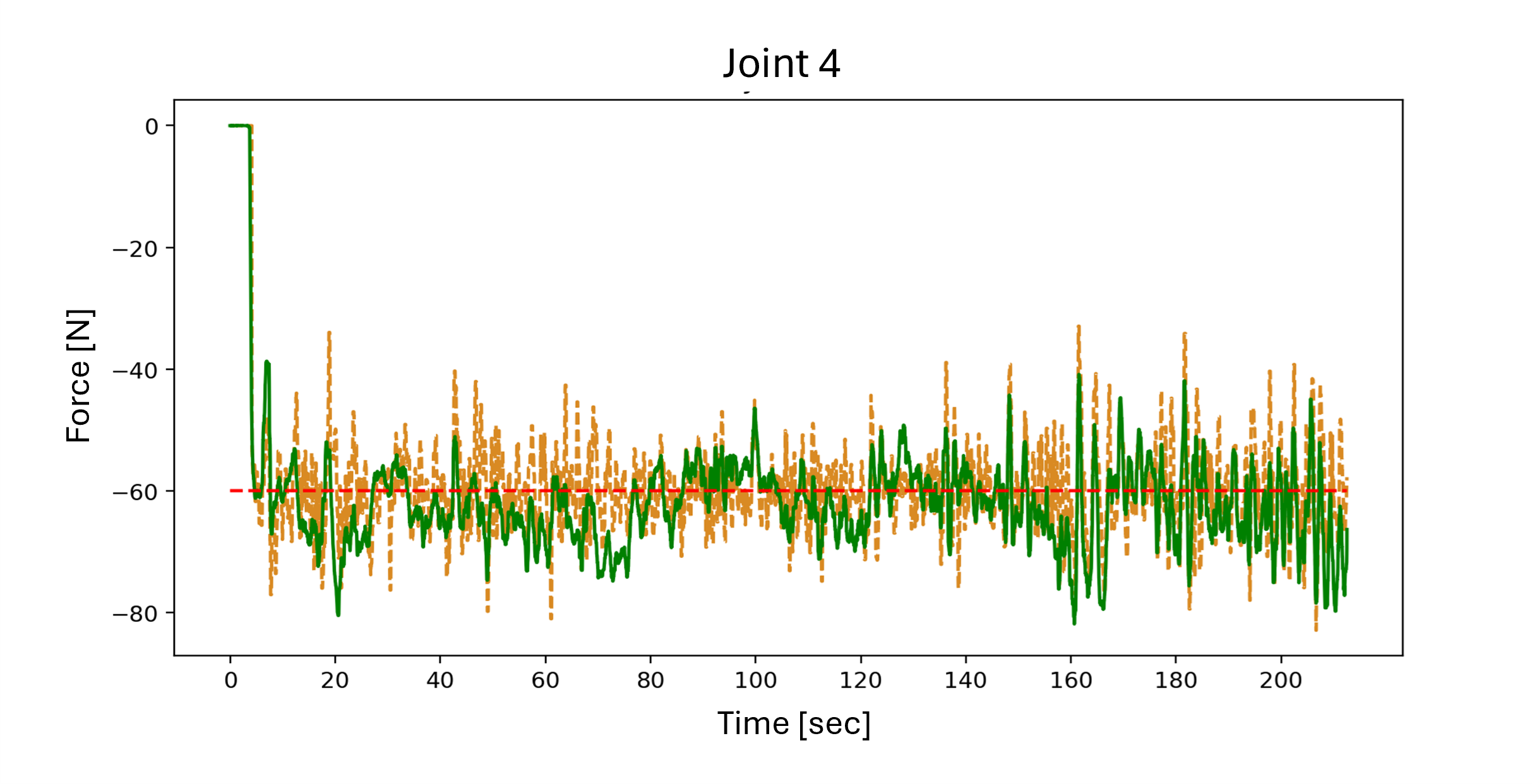}
    \caption{In the wiping the table experiment, the manipulator was set to wipe the table with constant 60 N force (red dashed line). The orange curve and green curve represent the estimated and measured contact forces between the robot and the table, respectively.}
    \label{fig: 12}
\end{figure}

The robot was set to move on a flat surface according to the trajectory, as shown in Figure~\ref{fig: 11}(a). The same commercial 6-axis force sensor was also mounted on the end effector to provide the ground truth force data as shown in Figure~\ref{fig: 11}(b). The robot was set to apply 60 N on the surface, and Figure~\ref{fig: 12} shows the results. The average error and standard deviation of the errors between the estimated and measured forces are 7.995 N and 5.843 N, respectively. Wiping the table was the first set of experiments conducted to evaluate the performance of the virtual force sensor. Unlike the peg-in-hole task, which utilized impedance control, simply digitized position control was utilized in this set of experiments, where the compensated displacement was directly determined by the force error without using impedance.

\subsection{Peg-in-hole}\label{subsec62}

The virtual force sensor was then applied in the peg-in-hole manipulation task, which required force/torque sensing information. As stated in the literature~\cite{c18,c19,c20,c23}, the peg-in-hole task was designed to be utilizable in the following circumstances: (i) no requirements for the models of the peg, the hole, or the environment; (ii) the peg and hole should be nearby initially, but their relative position and orientation are unknown; (iii) the hole does not have a guiding edge or a chamfer for easy assembly. 

\begin{figure}[http]
    \centering
    \includegraphics[width=320pt]{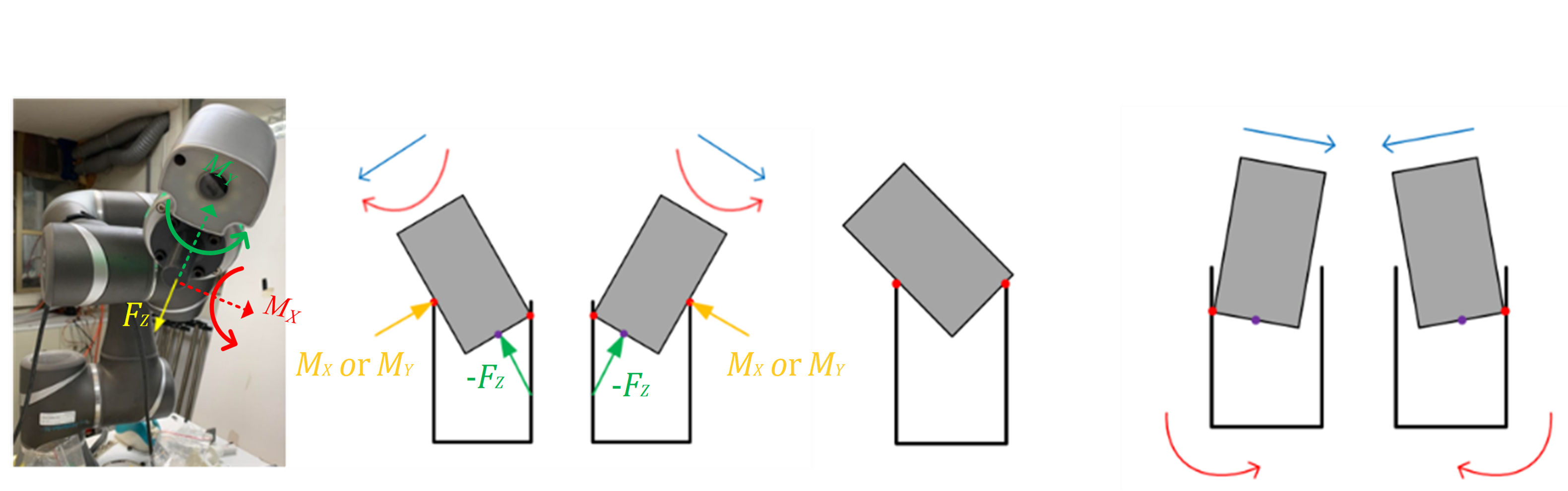}
    \caption{The peg-in-hole process: (a) approach the hole and two contact scenarios, (b) stuck outside the hole and (c) stuck inside the hole.}
    \label{fig: 13}
\end{figure}

\begin{figure}[http]
    \centering
    \includegraphics[width=320pt]{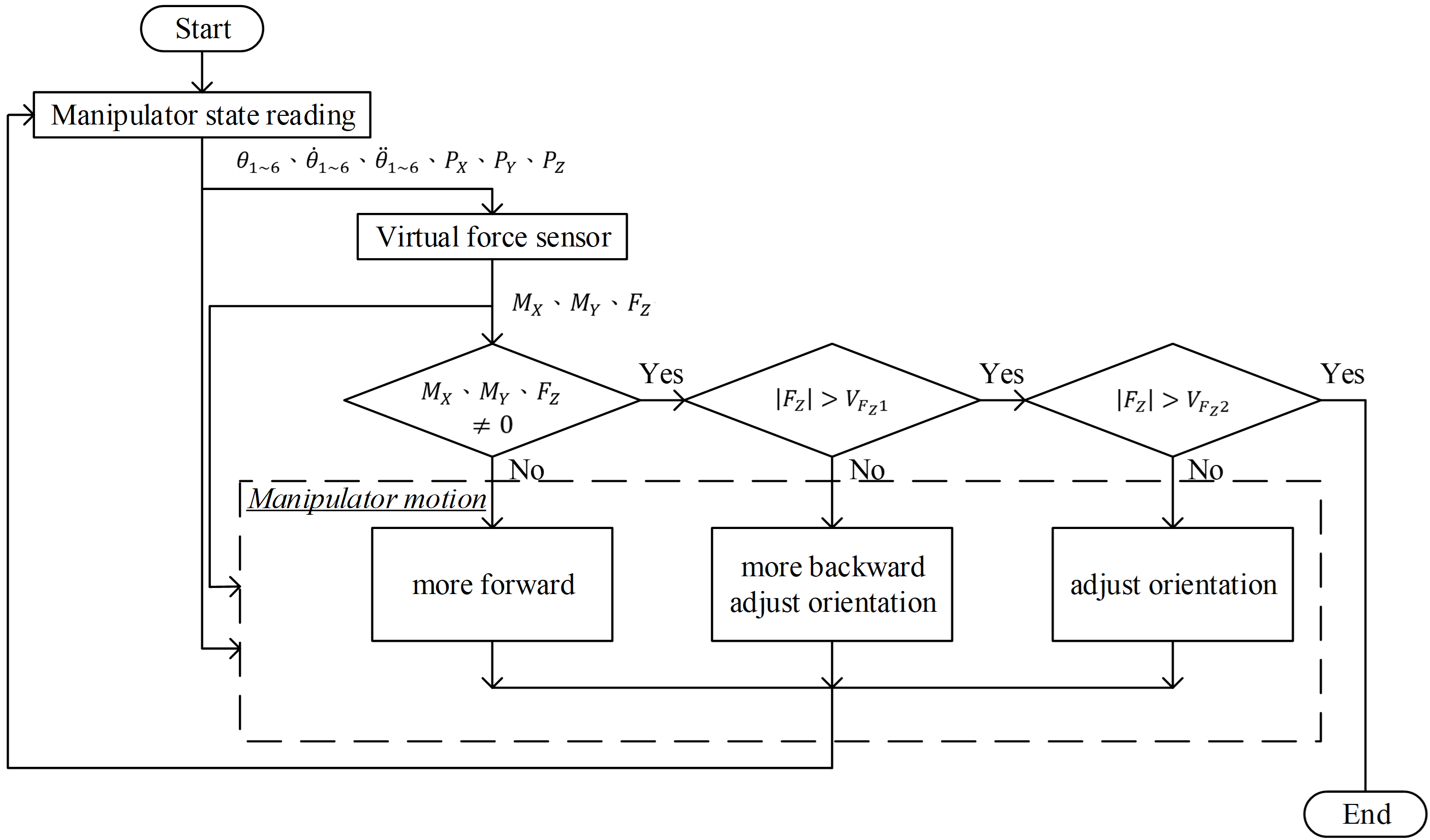}
    \caption{The control strategy of the peg-in-hole task; (a) the overall control structure; (b) the control flow chart.}
    \label{fig: 14}
\end{figure}

The peg-in-hole process involves multiple contact scenarios where the force/torque conditions vary. If the force/torque of the virtual sensor $(M_X, M_Y, F_Z)$ does not sense contact during the approach process, the peg keeps moving forward as shown in Figure~\ref{fig: 13}(a). The peg contacts the hole, two contact scenarios exist as shown in Figure~\ref{fig: 13}(b) and (c). When a peg is stuck in the hole shown in the small picture on the left of Figure~\ref{fig: 13}(b) and cannot get in, or when one side and bottom are stuck as shown in the small picture on the other side, the above two situations are regarded as "stuck outside the hole". In this case, the manipulator moves backward for a small distance and adjusts the posture simultaneously, where the peg rotates and moves laterally (i.e., red arrow and blue arrow). If the peg just contacts the hole with one side as shown in Figure~\ref{fig: 13}(c), this scenario is regarded as “stuck inside the hole,” and the peg is strategically rotated and moved laterally to eliminate this situation (i.e., red arrow and blue arrow). The overall control strategy of the peg in the hole is presented in Figure~\ref{fig: 14}.

During the peg-in-hole process, impedance control was utilized~\cite {c46, c47}. The system was modeled as a spring-damper mass system:
\begin{align}
m \ddot \theta_{dc} + b \dot \theta_{dc} + k(t)\theta_{dc} = \tau
\label{eqn: 22}
\end{align}
where $m,c,k,\tau$ represent mass, damping, spring, and external torque, and the state $\theta_{dc} = \theta_d - \theta_c$ describes the difference between the desired angle $\theta_d$ and the current angle $\theta_c$. To yield a smoother response, $k(t)$ was designed to be responsive to the torque difference~\cite{c46}, $e_\tau = \tau_d - \tau$:
\begin{align}
k(t) = k_{\tau}e_{\tau}{\theta_{dc}}^{-1} + k_v \dot e_{\tau} {\theta_{dc}}^{-1}
\label{eqn: 23}
\end{align}
After discretization and the rearrangement of terms, $\theta_{dc}$ could be represented as
\begin{align}
\theta_{dc} = (k_{\tau}\tau + k_v \frac{\tau}{\Delta t})/(\frac{m}{{\Delta t}^2} + \frac{b}{\Delta t})
\label{eqn: 24}
\end{align}
Then, the parameters $(m,b,k_{\tau},k_v,\Delta t)$ were selected based on the empirical manipulator.

After understanding the main three contact states of the peg and understanding how to convert the force change into position control (impedance controller), then look at the detailed description of the control strategy of the peg-in-hole task in Figure~\ref{fig: 14}. When starting to execute the strategy loop of the control strategy of the peg-in-hole task, the state of the manipulator (including the angular position of each joint, angular velocity, angular acceleration $\theta_{1-6}$, $\dot \theta_{1-6}$, $\ddot \theta_{1-6}$ will be read from the manipulator first, and the spatial coordinates $P_X, P_Y, P_Z$  of the end-effect), of which $\theta_{1-6}$, $\dot \theta_{1-6}$, $\ddot \theta_{1-6}$ are sent to the virtual force sensor gets $M_X, M_Y, F_Z$ 3-axis force information. Then, according to the 3-axis force information, it is determined which stage of the relationship between peg and hole is in Figure~\ref{fig: 13}, and determines which "Manipulator motion" command should be sent to the manipulator. (In addition to the judgment of the relationship between peg and hole at this stage, it is also necessary to obtain the current arm position $\theta_{1-6}$, $\dot \theta_{1-6}$, $\ddot \theta_{1-6}$, $P_X, P_Y, P_Z$ and end-effect 3-axis force information to determine how to adjust the posture and how the force should be applied).

\begin{figure}[http]
    \centering
    \includegraphics[width=360pt]{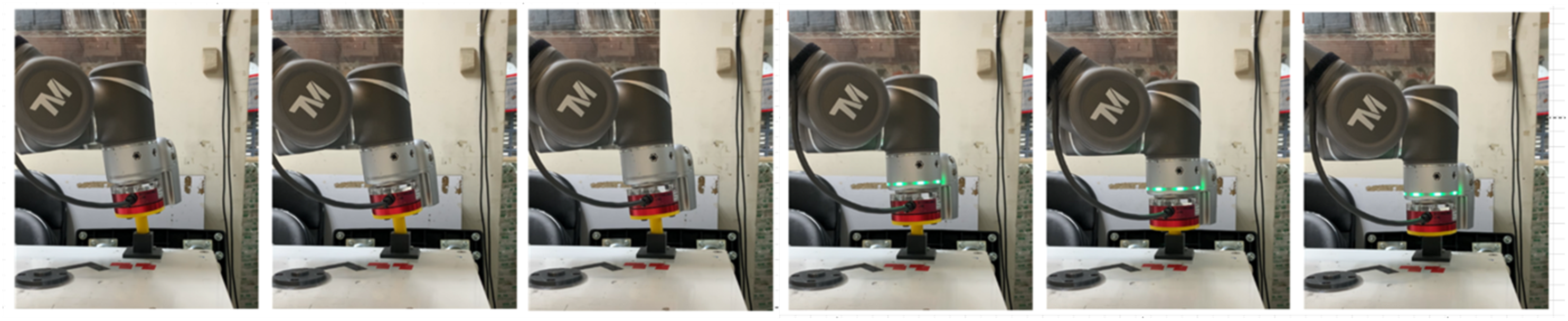}
    \caption{Snapshots of the peg-in-hole experiment.}
    \label{fig: 15}
\end{figure}

\begin{figure}[http]
    \centering
    \includegraphics[width=300pt]{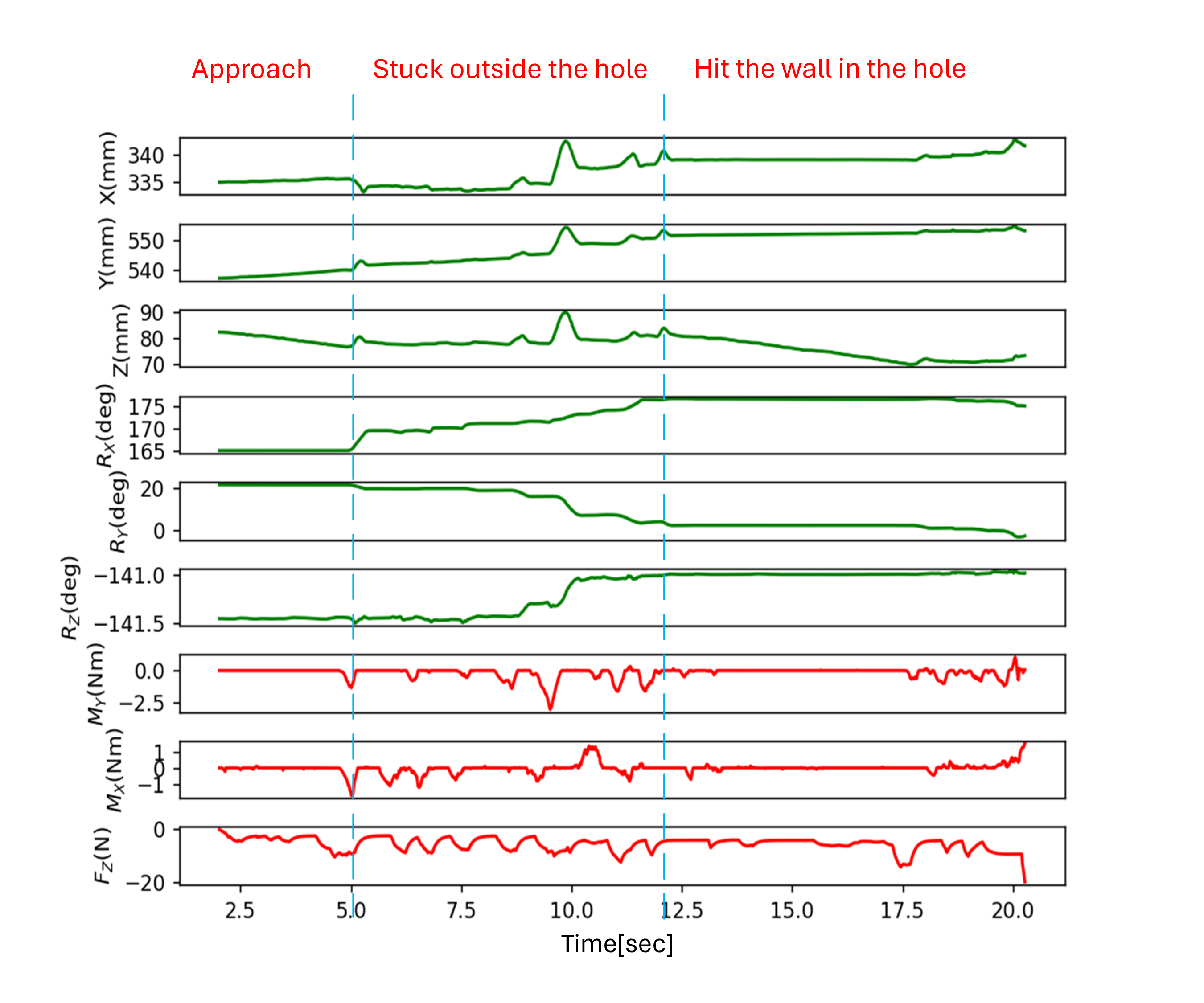}
    \caption{The green line during the Peg-in-hole mission is the displacement and direction of the peg. (The coordinate information is defined in Figure~\ref{fig: 1}(b)) The red is the force signal of the peg (the coordinates are defined in Figure~\ref{fig: 6}(a)).}
    \label{fig: 16}
\end{figure}

The peg-in-hole strategy, which utilizes information derived from the virtual force sensor was experimentally evaluated. The depth and diameter of the hole were set to 30 mm and 21.8 mm, respectively. Figure~\ref{fig: 15} shows snapshots of the experiments using a peg 40 mm in length and 220 mm in diameter. The mean absolute error (MAE) of the force/torque values between the commercial sensor and virtual sensors is 0.378 Nm $(M_X)$, 0.242 Nm $(M_Y)$, and 9.438 N $(F_Z)$, respectively. Figure~\ref{fig: 16} shows the states of the manipulator during the peg-in-hole process using the more tight-fit peg with a length of 48 mm and diameter of 21.4 mm. The time sequence shows that the peg approached the hole first, stuck there (i.e., a spike at 10 secs), and then an adjustment was executed to alter its position and orientation (as Figure~\ref{fig: 13}(b)) until the peg could successfully move forward until fully inserted. In the empirical system, real-time performance is mainly determined by the communication speed between the manipulator and PC, as well as how long it takes for the model to predict the output. The former is 125Hz, and the latter is about 63Hz.

\subsection{Speed optimization of trajectories}\label{subsec63}

The optimized trajectories were first learned and simulated with the physics-data hybrid dynamic model (simulator) and then the experiment was conducted with the TM5-700 manipulator to verify the trajectory planner. The baseline trajectory was generated by selecting three trajectories as Table~\ref{tab: Table 8} (Trajectory 1, 2, and 3). Each trajectory was planned with the simulator and the experimental results are in Table~\ref{tab: Table 8}. The 'Original' represents the elapsed time without speed optimization and the 'Optimized' stands for the time with speed optimization. The speed optimization method successfully reduced the elapsed time by an average $20 \%$ for three trajectories. To conclude, the motion planner developed in Section~\ref{sec5} is functional and succeeds in reducing the elapsed time of two selected trajectories.

\begin{table}[http]
\centering
\caption{\label{tab: Table 8} Trajectories before and after speed optimization. (Unit$\colon sec)$.}
    \begin{tabular}{|c c c c|} 
    \hline
    Time & Trajectory 1 & Trajectory 2 & Trajectory 3 \\ [0.5ex]
    \hline
    Original & 1.54 & 1.31 & 1.64\\
    Optimized & 1.16 & 1.11 & 1.29\\
    Improvement & 24.7 $\%$ & 15.3 $\%$ & 21.3 $\%$\\
    \hline
    \end{tabular}
\end{table}

\section{Conclusion}\label{sec7}
In this paper, we reported on the development of a virtual force sensor and a motion planner of a manipulator based on the physics-data hybrid dynamic model. The hybrid model has the best accuracy compared to the physics-based model and requires less training data compared to the data-driven model. Furthermore, the modeling results reveal that among the tested DNN-based, LSTM, and XGBoost architecture with hyperparameter optimization, XGBoost performs the most accurate modeling of the manipulator dynamics including un-modeled dynamics.

The external torque of the manipulator is then derived by subtracting the derived internal torque from the total motor torque. The external torque is further transformed into a 3-axis virtual force/torque on the end effector through a geometrical relation and machine learning technique. Finally, the virtual sensor is utilized in two applications. For the wiping the table task with a designated normal force of 60 N, the average error and standard deviation of the errors between the estimated and measured forces are 7.995 N and 5.843 N, respectively. For the peg-in-hole tasks, a peg 48 mm in length and 21.4 mm in diameter is able to be plugged into a hole 30 mm in depth and 21.8 mm in diameter. The MAE of the force/torque values between the commercial sensor and the virtual sensors are 0.378 Nm $(M_X)$, 0.242 Nm $(M_Y)$, and 9.438 N $(F_Z)$, respectively. Lastly, the learning-based motion planner successfully plans time-efficient trajectories for the manipulator. Three trajectories are tested and their elapsed time is reduced by an average 20.4 $\%$. The research results imply the potential to be implemented in industrial production lines.

We are in the process of refining the model so that it can compensate for the dead zone or friction effector of the manipulator more accurately. The model will be evaluated using different states as well. Furthermore, we plan to design an advanced controller and sim-to-real skill transfer for the manipulator based on the developed hybrid dynamic model (digital twin).

\clearpage

\bibliographystyle{asmems4}

\bibliography{asme2e}

\newpage 
\section*{Statements and Declarations}
\hfill
\subsection*{Funding}
\noindent This work is supported by the National Science and Technology Council (NSTC), Taiwan, under contract: MOST 110-2634-F-007-027- and MOST 111-2634-F-007-010-.

\subsection*{Competing Interests}
\noindent The authors have no relevant financial or non-financial interests to disclose.

\subsection*{Author Contributions}
\noindent All authors contributed to the study conception and design. Material preparation, data collection and analysis were performed by Jyun-Ming Liao and Wu-Te Yang. The first draft of the manuscript was written by Jyun-Ming Liao, and it was revised by Wu-Te Yang. The final manuscript was written by Pei-Chun Lin. Pei-Chun Lin also acquires funding and supervises and manages the project. All authors read and approved the final manuscript.

\subsection*{Data Availability}
\noindent The datasets generated during and/or analyzed during the current study are available from the corresponding author on reasonable request.


\end{document}